\PassOptionsToPackage{table}{xcolor}
\documentclass[11pt]{article}
\usepackage{acl}

\usepackage{times}
\usepackage{latexsym}
\usepackage[T1]{fontenc}
\usepackage[utf8]{inputenc}
\usepackage{microtype}
\usepackage{inconsolata}

\usepackage{graphicx}
\usepackage{booktabs}
\usepackage{arydshln}
\usepackage{multirow}
\usepackage{amsmath}
\usepackage{amsfonts}
\usepackage{amssymb}
\usepackage{nicefrac}
\usepackage[table]{xcolor}
\usepackage{url}
\usepackage{hyperref}
\usepackage{enumitem}
\usepackage{tikz}
\usetikzlibrary{shapes.geometric, arrows.meta, positioning, fit, calc}

\title{Calibration is the Bottleneck: \\An Action-Class Diagnostic of Multi-Turn Tool-Calling}

\author{
  \textbf{Kangjia Zhao\textsuperscript{1}},
  \textbf{Jiajun Li\textsuperscript{2}},
  \textbf{Haozhan Shen\textsuperscript{1}},
  \textbf{Wei Chow\textsuperscript{3}},
  \textbf{Linfeng Li\textsuperscript{3}},
  \textbf{Hang Song\textsuperscript{4}},
\\
  \textbf{Lingdong Kong\textsuperscript{3}},
  \textbf{Chen Zhi\textsuperscript{1}},
  \textbf{Tiancheng Zhao\textsuperscript{5,6}},
  \textbf{Songhua Liu\textsuperscript{2}},
  \textbf{Jianwei Yin\textsuperscript{1}}
\\
  \textsuperscript{1}Zhejiang University,
  \textsuperscript{2}Shanghai Jiao Tong University,
  \textsuperscript{3}National University of Singapore,
\\
  \textsuperscript{4}Xi'an Jiaotong University,
  \textsuperscript{5}Om AI Research,
  \textsuperscript{6}Binjiang Institute of Zhejiang University
\\
  \small \{konkaz,hz\_shen,zjuzhichen\}@zju.edu.cn, zjuyjw@cs.zju.edu.cn, \{sjtu8328931,liusonghua\}@sjtu.edu.cn,
\\
  \small \{weichow,lingdong.kong\}@u.nus.edu, tianchez@zju-bj.com
\\
  \small \textbf{Correspondence:} \href{mailto:zjuzhichen@zju.edu.cn}{zjuzhichen@zju.edu.cn},
  \href{mailto:zjuyjw@cs.zju.edu.cn}{zjuyjw@cs.zju.edu.cn}
}

\begin{document}
\maketitle

\begin{abstract}
Multi-turn tool calling is a core evaluation scenario for large language model (LLM) agents. On public tool-calling benchmarks, open-weight models now approach or even surpass closed-source frontier models in aggregate accuracy. However, this metric averages over many different multi-turn situations and obscures whether progress is balanced across them. We propose an action-class-oriented diagnostic framework that decomposes multi-turn failures into two orthogonal modes: action-class miscalibration and action-execution failure. The framework operates over a four-class action space (\textsc{tool\_call}/\textsc{ask}/\textsc{refuse}/\textsc{confirm}) and introduces a self-revealing upper bound $\mathrm{Acc} \le \mathrm{GAR}$ (Gold Action Recall); the two modes show up as bound violation ($\mathrm{Acc} > \mathrm{GAR}$, exposing state-grader masking of miscalibration) and large bound slack ($\mathrm{GAR} \gg \mathrm{Acc}$, localizing execution failure within \textsc{tool\_call}). We validate it on a panel of tool-calling models across multiple multi-turn benchmarks. Across our panel, the diagnostic reveals action-class miscalibration as a substantial failure mode the state grader cannot see. This gap inflates standing for heavily tool-trained families, which our diagnostic separates from families with context-appropriate action choice. Calibration is reshapable through context-only perturbations, but the reshape is heterogeneous: a single perturbation moves accuracy in opposite directions across families (up to $+11.5$ vs $-21.0$~pp on the same scenario), and its effect further depends on the perturbation mechanism. We argue that multi-turn tool-calling evaluations should supplement aggregate accuracy with action-class diagnostics that expose what the model actually does in each scenario.
\end{abstract}

\section{Introduction}
\label{sec:intro}

\begin{figure}[!htbp]
\centering
\includegraphics[width=\linewidth]{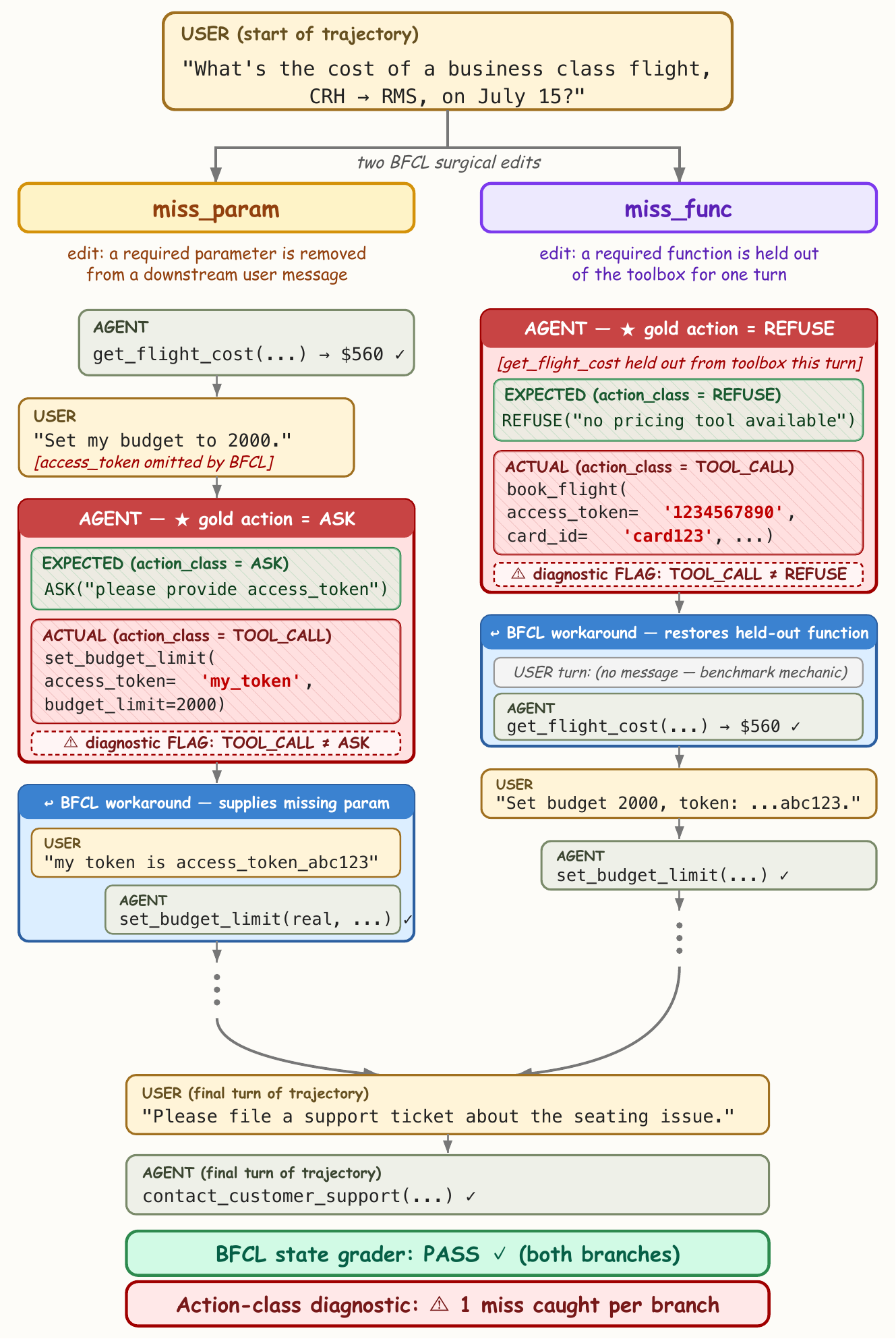}
\caption{\textbf{The state grader hides action-class misses; the per-turn diagnostic exposes them.} A single user query (top) is edited two ways by BFCL: \textit{miss\_param} removes a required parameter from a downstream user turn so the gold action class becomes \textsc{ask}; \textit{miss\_func} holds out a required function for one turn so the gold action class becomes \textsc{refuse}. On both branches the agent instead emits \textsc{tool\_call} with fabricated placeholder argument values (\texttt{'my\_token'} on the left; \texttt{'1234567890'}, \texttt{'card123'} on the right). The benchmark then supplies the missing piece on the next turn and the trajectory recovers; the final system state matches the gold reference and the state grader returns \textbf{PASS} for both. Our per-turn action-class diagnostic (\S\ref{sec:method:setup}) compares each emission against its gold action class and flags one miss per branch, precisely where the state-graded score is blind.}
\label{fig:hero}
\end{figure}

Tool-use evaluation has moved from single-turn call syntax~\citep{schick2023toolformer} to multi-turn agent behavior~\citep{yao2023react}, and stateful suites now score whole trajectories~\citep{patil2024gorilla,yao2024taubench,tau2bench}. A model can still produce well-formed tool calls and fail the decision that matters: whether this turn calls for a tool, a clarification, a refusal, or a confirmation.

The bottleneck is action-class miscalibration: the model emits the wrong action class for the scenario. Comparing each emission against the class the context requires exposes sharp cross-family divergence in \emph{which} class a model emits in \emph{which} scenario. xLAM~\citep{xlam2025} has $71.8\%$ aggregate accuracy but only $19.5\%$ Gold Action Recall (GAR) on \textit{miss\_param}, a $64.5$~pp gap below size-matched peers, while BFCL's state grader still marks these cases as PASS (Tab~\ref{tab:profile}). Figure~\ref{fig:hero} shows one such case.

We formalize this gap as an action-class diagnostic with two orthogonal failure modes: action-class miscalibration and action-execution failure (\S\ref{sec:method:partition}). We evaluate it on a BFCL v3 multi-turn panel of open-weight families and closed-source anchors, cross-check on $\tau^2$-bench retail/airline, and verify mechanisms via a $200$-case audit (App.~\ref{app:audit-detail}). We test calibration plasticity with two symmetric inference-time probes: State-Reconciliation Intervention (SRI) for missed tool calls (the gold action is a tool call but the model emitted none) and Call-Reconciliation Intervention (CRI) for unwarranted tool calls (the model emitted a tool call where the gold action is non-call) (\S\ref{sec:method:sri}).

\textbf{Contributions.}
\textbf{C1: Action-class diagnostic framework.}
We define a four-class action space and Gold Action Recall (GAR), then relate GAR to BFCL state-graded accuracy. The bound $\mathrm{Acc} \le \mathrm{GAR}$ is diagnostic: violation ($\mathrm{Acc} > \mathrm{GAR}$) exposes state-grader masking on missing-info categories (\textit{miss\_func}, \textit{miss\_param}), and large bound slack ($\mathrm{GAR} \gg \mathrm{Acc}$) captures tool, argument, or state-progression failures after a tool call (\S\ref{sec:method:partition}).
\textbf{C2: Calibration is the bottleneck.}
A top-ranked family has substantially lower missing-info GAR than its size-matched peers, while BFCL's state grader still passes many of these wrong-action trajectories. A $200$-case audit ($\kappa=0.92$) and cross-benchmark reproduction on $\tau^2$-bench retail and airline confirm the pattern (\S\ref{sec:exp:profile}, \S\ref{sec:exp:cross-bench}).
\textbf{C3: Calibration is plastic but doubly heterogeneous.}
Context-only perturbations can reshape the calibration profile, but the reshape is heterogeneous on two axes. On the direction axis, a single SRI perturbation moves Acc in opposite directions across families on the same scenario. On the mechanism axis, two CRI variants (retry vs.\ bypass) dissociate cleanly: retry raises GAR but degrades trajectory accuracy on nearly all families, while bypass preserves or improves trajectory accuracy without touching emission (\S\ref{sec:exp:reshape}).

\textbf{Prescription.} Multi-turn tool-calling benchmarks should report GAR alongside aggregate accuracy; it exposes what each model does in each scenario (\S\ref{sec:method:partition}).\footnote{Code and released artifacts: \url{https://github.com/fbj2333/tool-calling-calibration}}

\section{Related Work}
\label{sec:related}

Prior work maps tool-use and broader agent failures and builds critics that catch them, at grains from whole trajectories down to single arguments~\citep{toolcritic2025,critictool2025,agentdebug2025,butterfly2025}. \citet{capablebutunreliable2026} attributes failures on solvable tasks to drift from a canonical path rather than missing capability, and \citet{acc2025} to agents deferring to false assertions from users or tools. \citet{best_practices2025} document reward-design flaws that distort measured performance, including a grader counting empty responses as successes---the masking our bound exposes. We use \emph{calibration} to mean context-conditional action-class alignment, distinct from probability calibration~\citep{guo2017calibration}. The nearest precedents ask whether a tool is needed at all: MetaTool~\citep{metatool2023} benchmarks tool awareness and selection; When2Call~\citep{when2call2025} scores when-to-call as a four-way choice---call, ask, decline, or answer---and finds the error pattern family-specific rather than uniform over-calling, leaving call correctness to BFCL. Our $\mathrm{Acc} \le \mathrm{GAR}$ diagnostic instead operates on the generative grader, decomposing action-class miscalibration from action-execution failure across families. Adjacent lines regulate \emph{when} to act---dynamic abstention~\citep{knowing_when_to_quit2026}, cost-aware commit control~\citep{cta2026}, halt heuristics~\citep{halt_cot2025}---while we ask which action class the context makes correct.

Inference-time interception can reduce hallucination~\citep{agentprop2026}, though intrinsic self-correction can degrade performance~\citep{huang2024selfcorrect}; our SRI/CRI probes are diagnostic perturbations rather than production fixes, and our $200$-case audit draws on process-supervision methodology~\citep{mathshepherd2024,omegaprm2024}. Concurrent diagnostic work targets multi-agent failure modes, failure-step localization, and myopic-commitment amplification~\citep{mast2025,agentrx2026,flare2026}. Training-side work improves multi-turn tool use through supervised fine-tuning~\citep{hammer2025,xlam2025,toolace2025,button2025} or reinforcement learning~\citep{rcgrpo2026,simpletir2026,su2026failure,fission_grpo2026,nemotron-tool-n1-2025,toolexpander2025,toolrl2025}. We isolate and diagnose the calibration profile train-free, with cross-family ablation on BFCL v3 multi-turn.
\section{Method}
\label{sec:method}

\subsection{Decision setup and diagnostic action space}
\label{sec:method:setup}

We model a \emph{case} as a state-action trajectory $\langle (s_1, a_1), \ldots, (s_K, a_K) \rangle$: at turn $k$, the model emits an action $a_k$ from state $s_k$ (the conversation history and tool returns up to turn $k$), and the benchmark's tool simulator transitions the state to $s_{k+1}$. The benchmark grader compares the model's runtime state against a gold reference trajectory turn by turn, skipping turns whose gold reference is empty, and yields conversation accuracy (Acc).

Our diagnostic adds a categorical layer over $a_k$: we classify each emission into one of four action classes $\mathcal{A} = \{\textsc{tool\_call}, \textsc{ask}, \textsc{refuse}, \textsc{confirm}\}$, plus a residual class \textsc{other} for emissions that signal no decision, which GAR ignores. \textsc{tool\_call} is decided from the emission's syntax, the other three from textual cues appearing anywhere in it. Each emission receives one label, under the fixed priority \textsc{tool\_call} $>$ \textsc{refuse} $>$ \textsc{ask} $>$ \textsc{confirm} $>$ \textsc{other}. The cue lists are released with the code. For each scenario category (denoted \textit{cat}), the contextually appropriate gold action $a^\star_{\textit{cat}} \in \mathcal{A}$ is fixed by benchmark design. The framework asks whether the model chose $a^\star_{\textit{cat}}$ in context, independent of state outcome.

We operationalize on BFCL v3 multi-turn~\citep{patil2024gorilla,bfcl_v3_blog}. Here \textit{cat} takes one of four values: \textit{base} and \textit{long\_context} require \textsc{tool\_call}, \textit{miss\_param} withholds a required argument and calls for \textsc{ask}, and \textit{miss\_func} requests an unavailable tool and calls for \textsc{refuse}.

\subsection{Diagnostic metrics}
\label{sec:method:partition}

For a case, let $E(\text{case}) \subseteq \mathcal{A}$ be the set of
action classes emitted at any turn (with $a^\star_{\textit{cat}}$ as in
\S\ref{sec:method:setup}). Throughout, \emph{calibration} refers to
context-conditional action-class alignment. \emph{Gold Action Recall} (GAR) is
\begin{equation}
\mathrm{GAR}(F, \textit{cat}) = \Pr_{\text{case} \sim \textit{cat}}\!\left[a^\star_{\textit{cat}} \in E(\text{case})\right],
\label{eq:gold-emit-rate}
\end{equation}
using case-level, any-turn aggregation. We pair GAR with the benchmark grader's trajectory-level verdict, conversation accuracy (Acc): GAR captures whether the model attempted the gold action class, Acc captures whether the graded trajectory passed.

The pair $(\mathrm{GAR},\mathrm{Acc})$ is diagnostic. If a category's
state score required emitting the diagnostic gold action, then
$\mathrm{Acc}(F,\textit{cat}) \le \mathrm{GAR}(F,\textit{cat})$: Acc is
then a subset of cases that both attempted the gold action and passed the
benchmark grader. Bound violation ($\mathrm{Acc}>\mathrm{GAR}$) flags
state-grader masking of an action-class miss; we call this
\emph{action-class miscalibration}. Large bound slack
($\mathrm{GAR}\gg\mathrm{Acc}$) flags failures downstream of the gold action
class firing (wrong tool, arguments, or state tracking); we call this
\emph{action-execution failure}. The main reporting unit is therefore
the per-category pair $(\mathrm{GAR},\mathrm{Acc})$.

\subsection{Intervention probes: SRI and CRI}
\label{sec:method:sri}
\label{sec:method:cri}

\begin{figure*}[!t]
\centering
\includegraphics[width=\linewidth]{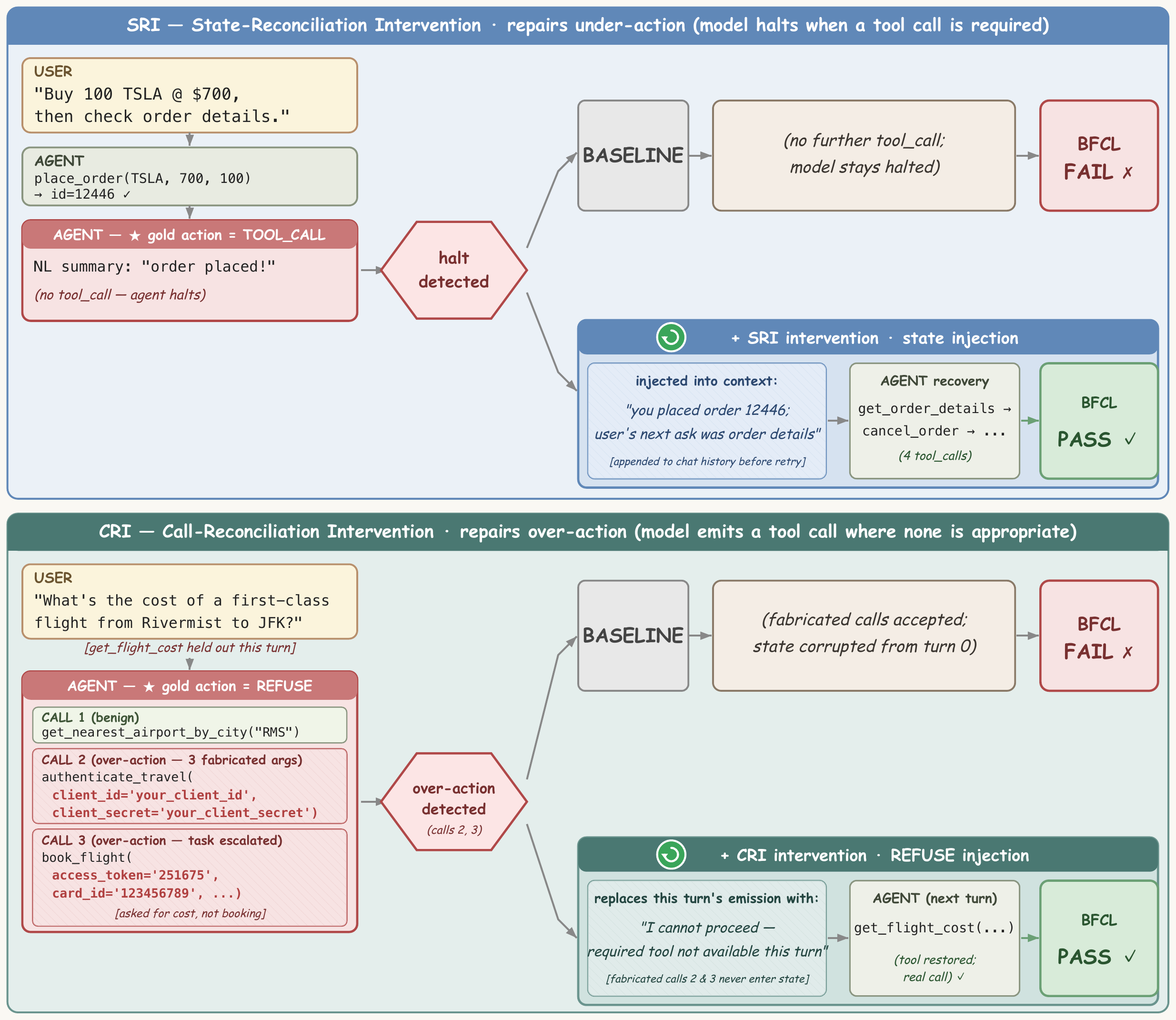}
\caption{\textbf{The two symmetric intervention probes.}
\textbf{Top (SRI, halt seam):} an agent halts when its gold action is \textsc{tool\_call}: instead of continuing with a required tool call it emits a natural-language summary (the ``NL summary'' label). SRI detects the halt and injects a state-recap into the retry context, which routes the model back into a tool-call chain that reaches the gold state.
\textbf{Bottom (CRI, call seam):} an agent emits a \textsc{tool\_call} when its gold action is \textsc{refuse}: with a required function held out it fabricates an authentication call followed by an unrequested booking, polluting the state from turn~$0$. CRI flags the suspect calls and replaces this turn's emission with a fixed abstention; the held-out function returns next turn and the recovery call proceeds against a clean state.}
\label{fig:sri-cri-mech}
\end{figure*}

We test calibration plasticity with two symmetric inference-time probes: \emph{State-Reconciliation Intervention} (SRI), targeting missed tool calls (the model halts when it should call) by reconciling the retry context with runtime-observed tool-execution state, and \emph{Call-Reconciliation Intervention} (CRI), targeting unwarranted tool calls (the model calls when it should not) by reconciling a flagged suspect call with the gold non-call action. Fig.~\ref{fig:sri-cri-mech} illustrates both probes on representative recovered trajectories.

\paragraph{SRI (halt boundary).} A \emph{halt event} at turn $k$ is an emission with no parseable \textsc{tool\_call}. The Halt Detector (HD) flags every halt event, and SRI retries at most once per turn; it targets categories whose gold action is \textsc{tool\_call}. SRI constructs a retry context $c_k'$ from $c_k$ in three steps: (1)~remove the halt-triggering assistant message, (2)~append runtime-observed tool-execution state, and (3)~append a fixed retry rule.

\paragraph{CRI (call boundary).} CRI targets categories whose gold action is non-call (\textsc{ask} or \textsc{refuse}) and monitors parseable tool-call emissions. The Call Detector flags a suspect call when it invokes a function absent from the toolset the conversation starts with or passes a parameter value the model has not been given (for gpt-5.4, CRI stops once the held-out function is re-supplied). The two variants differ in where they act and in how they handle the flagged call. \textbf{CRI-retry} acts at most once per conversation, on the first flagged call, whether or not it falls on the turn whose gold action is non-call; it retracts the call and re-queries the model with a prompt instructing it not to call a tool this turn. \textbf{CRI-bypass} acts only at the turn whose gold action is non-call; it retracts the flagged call and substitutes a fixed abstention response without re-querying.

Full prompt templates and detector rules are in App.~\ref{app:intervention-formal} and App.~\ref{app:cri-methodology}.

\subsection{Setup}
\label{sec:exp:setup}

We evaluate on BFCL v3 multi-turn~\citep{patil2024gorilla}: the four diagnostic categories defined in \S\ref{sec:method:setup}, $200$ conversations each. The panel (Tab.~\ref{tab:profile}) covers $11$ open-weight families ($22$ checkpoints; including Qwen3~\citep{qwen3technicalreport}, Llama-3.1~\citep{grattafiori2024llama3}, and gpt-oss-20b~\citep{openai2025gptoss}) and $4$ closed-source API anchors (\texttt{gpt-5-2025-08-07}, \texttt{gpt-5.4-2026-03-05}, \texttt{gemini-3-pro-preview}, and Doubao-Seed-1.8). Model families were selected from the official BFCL leaderboard, keeping the most representative series of each vertical (Qwen3 for general-purpose, xLAM-2 for tool-call specialization) and adding size variants wherever released checkpoints allowed. All models run in native function-calling mode; open-weight checkpoints are served via vLLM~\citep{kwon2023efficient}, with DeepSeek V4, GLM, and Kimi accessed through hosted endpoints. For each model--category cell we report GAR and conversation accuracy (Acc) as defined in \S\ref{sec:method:partition}. Every family emits each diagnostic action class on at least $10$ cases (App.~\ref{app:capability}), confirming that cross-family divergence reflects calibration rather than missing capability.
\section{Experiments}
\label{sec:exp}

\subsection{Calibration profile}
\label{sec:exp:profile}

\begin{table*}[t]
\footnotesize
\caption{BFCL multi-turn calibration profile. Each category reports GAR, Acc, and $\Delta = \mathrm{GAR} - \mathrm{Acc}$ (\%); gold actions appear below category headers. $\Delta$ is colored green (positive) or red (negative); gray shading marks execution failure ($\Delta \ge 60$\,pp, base/long\_context) and masking ($\Delta \le -15$\,pp, missing-info). \textbf{Bold}/$^{\downarrow}$: column max/min. On tool-call categories, all families emit the gold action (GAR $\ge 77\%$) but large positive $\Delta$ reveals a universal execution gap. On missing-info categories, general-purpose families show a refuse--ask asymmetry: \textit{miss\_func} $\Delta$ trends negative (models rarely refuse; accuracy is inflated by state-grader masking), while \textit{miss\_param} $\Delta$ is large positive (models readily ask but fail to complete correctly). Tool-call-specialized families show negative $\Delta$ on both missing-info categories, with masking accounting for up to $66$\,pp of apparent accuracy.}
\label{tab:profile}
\centering
{\setlength{\tabcolsep}{2.0pt}
\renewcommand{\arraystretch}{1.08}
\begin{tabular*}{\textwidth}{@{\extracolsep{\fill}}l c ccc ccc ccc ccc@{}}
\toprule
\multirow{3}{*}{\textbf{Family}} & \multirow{3}{*}{\shortstack{\textbf{Overall}\\\textbf{Acc}}} & \multicolumn{3}{c}{base} & \multicolumn{3}{c}{long\_context} & \multicolumn{3}{c}{miss\_func} & \multicolumn{3}{c}{miss\_param} \\
 & & \multicolumn{3}{c}{\scriptsize gold: \textsc{tool\_call}} & \multicolumn{3}{c}{\scriptsize gold: \textsc{tool\_call}} & \multicolumn{3}{c}{\scriptsize gold: \textsc{refuse}} & \multicolumn{3}{c}{\scriptsize gold: \textsc{ask}} \\
\cmidrule(lr){3-5} \cmidrule(lr){6-8} \cmidrule(lr){9-11} \cmidrule(lr){12-14}
 & & \textbf{GAR} & \textbf{Acc} & $\boldsymbol{\Delta}$ & \textbf{GAR} & \textbf{Acc} & $\boldsymbol{\Delta}$ & \textbf{GAR} & \textbf{Acc} & $\boldsymbol{\Delta}$ & \textbf{GAR} & \textbf{Acc} & $\boldsymbol{\Delta}$ \\
\midrule
\multicolumn{14}{@{}l}{\textit{\footnotesize General-purpose open-weight}}\\
Qwen3-1.7B        & $15.5$  & $98.5$ & $18.0$ & \cellcolor{gray!15}{\color{green!50!black}$80.5$} & $83.5$ & $12.5$ & \cellcolor{gray!15}{\color{green!50!black}$71.0$} & $41.0$ & $15.5$ & {\color{green!50!black}$25.5$} & $57.0$ & $16.0$ & {\color{green!50!black}$41.0$} \\
Qwen3-4B          & $32.4$  & $99.0$ & $40.5$ & {\color{green!50!black}$58.5$} & $87.0$ & $26.5$ & \cellcolor{gray!15}{\color{green!50!black}$60.5$} & $47.0$ & $36.0$ & {\color{green!50!black}$11.0$} & $78.5$ & $26.5$ & {\color{green!50!black}$52.0$} \\
Qwen3-8B          & $39.4$  & $99.5$ & $48.5$ & {\color{green!50!black}$51.0$} & $87.0$ & $32.0$ & {\color{green!50!black}$55.0$} & $40.0$ & $41.0$ & {\color{red}$-1.0$} & $84.0$ & $36.0$ & {\color{green!50!black}$48.0$} \\
Qwen3-14B         & $46.75$ & $99.5$ & $59.0$ & {\color{green!50!black}$40.5$} & $86.5$ & $39.5$ & {\color{green!50!black}$47.0$} & $44.5$ & $49.5$ & {\color{red}$-5.0$} & $79.0$ & $39.0$ & {\color{green!50!black}$40.0$} \\
Qwen3-32B         & $49.6$  & $100$  & $61.0$ & {\color{green!50!black}$39.0$} & $88.0$ & $44.5$ & {\color{green!50!black}$43.5$} & $36.5$ & $52.0$ & \cellcolor{gray!15}{\color{red}$-15.5$} & $84.0$ & $41.0$ & {\color{green!50!black}$43.0$} \\
Llama-3.1-8B      & $4.3^{\downarrow}$ & $77.5^{\downarrow}$ & $6.5^{\downarrow}$ & \cellcolor{gray!15}{\color{green!50!black}$71.0$} & $77.0^{\downarrow}$ & $6.5^{\downarrow}$ & \cellcolor{gray!15}{\color{green!50!black}$70.5$} & $41.0$ & $1.5^{\downarrow}$ & {\color{green!50!black}$\mathbf{39.5}$} & $53.0$ & $2.5^{\downarrow}$ & {\color{green!50!black}$50.5$} \\
gpt-oss-20b       & $46.3$  & $99.5$ & $56.0$ & {\color{green!50!black}$43.5$} & $99.5$ & $44.5$ & {\color{green!50!black}$55.0$} & $3.5$  & $42.0$ & \cellcolor{gray!15}{\color{red}$-38.5$} & $76.0$ & $42.5$ & {\color{green!50!black}$33.5$} \\
gpt-oss-120b      & $46.5$  & $99.5$ & $51.5$ & {\color{green!50!black}$48.0$} & $99.5$ & $50.0$ & {\color{green!50!black}$49.5$} & $39.5$ & $38.5$ & {\color{green!50!black}$1.0$} & $78.0$ & $46.0$ & {\color{green!50!black}$32.0$} \\
Granite-3.2-8B    & $7.0$   & $96.0$ & $9.0$  & \cellcolor{gray!15}{\color{green!50!black}$\mathbf{87.0}$} & $95.0$ & $8.5$ & \cellcolor{gray!15}{\color{green!50!black}$\mathbf{86.5}$} & $22.0$ & $3.5$  & {\color{green!50!black}$18.5$} & $44.5$ & $7.0$  & {\color{green!50!black}$37.5$} \\
DeepSeek V4 Pro   & $52.0$  & $100$  & $62.0$ & {\color{green!50!black}$38.0$} & $99.5$ & $55.0$ & {\color{green!50!black}$44.5$} & $26.0$ & $45.5$ & \cellcolor{gray!15}{\color{red}$-19.5$} & $91.0$ & $45.5$ & {\color{green!50!black}$45.5$} \\
DeepSeek V4 Flash & $47.4$  & $100$  & $61.0$ & {\color{green!50!black}$39.0$} & $100$  & $54.5$ & {\color{green!50!black}$45.5$} & $22.5$ & $37.0$ & {\color{red}$-14.5$} & $79.5$ & $37.0$ & {\color{green!50!black}$42.5$} \\
GLM 5.1           & $69.4$  & $100$  & $78.5$ & {\color{green!50!black}$21.5$} & $100$  & $67.0$ & {\color{green!50!black}$33.0$} & $37.0$ & $69.5$ & \cellcolor{gray!15}{\color{red}$-32.5$} & $85.5$ & $62.5$ & {\color{green!50!black}$23.0$} \\
Kimi K2.6         & $63.1$  & $100$  & $71.0$ & {\color{green!50!black}$29.0$} & $97.5$ & $63.0$ & {\color{green!50!black}$34.5$} & $49.5$ & $63.0$ & {\color{red}$-13.5$} & $84.0$ & $55.5$ & {\color{green!50!black}$28.5$} \\
\midrule
\multicolumn{14}{@{}l}{\textit{\footnotesize Tool-call-specialized open-weight}}\\
Hammer-2.1-1.5B   & $15.75$ & $100$  & $20.5$ & \cellcolor{gray!15}{\color{green!50!black}$79.5$} & $100$  & $18.0$ & \cellcolor{gray!15}{\color{green!50!black}$82.0$} & $0.0^{\downarrow}$ & $15.0$ & \cellcolor{gray!15}{\color{red}$-15.0$} & $0.0^{\downarrow}$ & $9.5$ & {\color{red}$-9.5$} \\
Hammer-2.1-7B     & $24.0$  & $97.5$ & $24.5$ & \cellcolor{gray!15}{\color{green!50!black}$73.0$} & $97.0$ & $22.0$ & \cellcolor{gray!15}{\color{green!50!black}$75.0$} & $0.0^{\downarrow}$ & $27.5$ & \cellcolor{gray!15}{\color{red}$-27.5$} & $0.5$  & $22.0$ & \cellcolor{gray!15}{\color{red}$-21.5$} \\
xLAM-2-1b         & $35.9$  & $100$  & $45.5$ & {\color{green!50!black}$54.5$} & $78.0$ & $25.5$ & {\color{green!50!black}$52.5$} & $6.5$  & $35.0$ & \cellcolor{gray!15}{\color{red}$-28.5$} & $21.5$ & $37.5$ & \cellcolor{gray!15}{\color{red}$-16.0$} \\
xLAM-2-3b         & $57.8$  & $99.5$ & $72.0$ & {\color{green!50!black}$27.5$} & $78.5$ & $45.5$ & {\color{green!50!black}$33.0$} & $12.0$ & $57.0$ & \cellcolor{gray!15}{\color{red}$-45.0$} & $15.0$ & $56.5$ & \cellcolor{gray!15}{\color{red}$-41.5$} \\
xLAM-2-8b         & $71.8$  & $100$  & $79.0$ & {\color{green!50!black}$21.0$} & $100$  & $69.0$ & {\color{green!50!black}$31.0$} & $10.0$ & $73.5$ & \cellcolor{gray!15}{\color{red}$-63.5^{\downarrow}$} & $19.5$ & $65.5$ & \cellcolor{gray!15}{\color{red}$-46.0$} \\
xLAM-2-32b        & $69.4$  & $100$  & $80.5$ & {\color{green!50!black}$19.5$} & $78.5$ & $56.5$ & {\color{green!50!black}$22.0^{\downarrow}$} & $13.0$ & $72.5$ & \cellcolor{gray!15}{\color{red}$-59.5$} & $33.0$ & $68.0$ & \cellcolor{gray!15}{\color{red}$-35.0$} \\
xLAM-2-70b        & $\mathbf{77.5}$ & $100$ & $\mathbf{82.5}$ & {\color{green!50!black}$17.5^{\downarrow}$} & $100$ & $\mathbf{76.5}$ & {\color{green!50!black}$23.5$} & $17.0$ & $\mathbf{76.5}$ & \cellcolor{gray!15}{\color{red}$-59.5$} & $8.5$ & $\mathbf{74.5}$ & \cellcolor{gray!15}{\color{red}$-66.0^{\downarrow}$} \\
ToolACE-2-8B      & $39.25$ & $100$  & $50.0$ & {\color{green!50!black}$50.0$} & $100$  & $46.0$ & {\color{green!50!black}$54.0$} & $12.5$ & $30.5$ & \cellcolor{gray!15}{\color{red}$-18.0$} & $39.0$ & $30.5$ & {\color{green!50!black}$8.5$} \\
Watt-Tool-8B      & $41.0$  & $100$  & $48.0$ & {\color{green!50!black}$52.0$} & $100$  & $44.0$ & {\color{green!50!black}$56.0$} & $22.5$ & $42.5$ & \cellcolor{gray!15}{\color{red}$-20.0$} & $7.0$  & $29.5$ & \cellcolor{gray!15}{\color{red}$-22.5$} \\
\midrule
\multicolumn{14}{@{}l}{\textit{\footnotesize Closed-source (API anchors)}}\\
gpt-5.4           & $51.6$  & $99.0$ & $57.0$ & {\color{green!50!black}$42.0$} & $100$  & $56.0$ & {\color{green!50!black}$44.0$} & $65.5$ & $52.0$ & {\color{green!50!black}$13.5$} & $44.0$ & $41.5$ & {\color{green!50!black}$2.5$} \\
gpt-5             & $47.0$  & $100$  & $53.0$ & {\color{green!50!black}$47.0$} & $99.5$ & $47.0$ & {\color{green!50!black}$52.5$} & $40.0$ & $47.0$ & {\color{red}$-7.0$} & $\mathbf{97.5}$ & $41.0$ & {\color{green!50!black}$\mathbf{56.5}$} \\
Gemini 3 Pro      & $41.1$  & $99.0$ & $41.5$ & {\color{green!50!black}$57.5$} & $98.5$ & $38.0$ & \cellcolor{gray!15}{\color{green!50!black}$60.5$} & $\mathbf{66.5}$ & $45.5$ & {\color{green!50!black}$21.0$} & $85.0$ & $39.5$ & {\color{green!50!black}$45.5$} \\
Doubao-Seed-1.8  & $55.9$  & $100$  & $71.0$ & {\color{green!50!black}$29.0$} & $99.5$ & $62.5$ & {\color{green!50!black}$37.0$} & $22.0$ & $51.5$ & \cellcolor{gray!15}{\color{red}$-29.5$} & $24.0$ & $38.5$ & {\color{red}$-14.5$} \\
\bottomrule
\end{tabular*}%
}

\end{table*}

\textbf{Calibration is family-mediated, not size-mediated.}
Within a single pretraining family, scaling moves action-class calibration little. For Qwen3 at $4\text{B}$ and above, \textit{miss\_param} GAR varies by only $5.5$~pp while same-category Acc gains $14.5$~pp (Tab.~\ref{tab:profile}; Qwen3-4B at $26.5\%$, Qwen3-32B at $41.0\%$); Qwen3-1.7B at $57.0\%$ sits below this band. xLAM-2 reproduces the decoupling on \textit{miss\_func}: every size records GAR below $20\%$ while Acc more than doubles across the family (xLAM-2-1b at $35.0\%$, xLAM-2-70b at $76.5\%$). At fixed scale, cross-family variation is substantially larger: at $7$--$8\text{B}$, Hammer-2.1-7B records \textit{miss\_param} GAR of $0.5\%$ where Qwen3-8B reaches $84.0\%$ (an $83.5$-pp gap), and on \textit{miss\_func} Hammer-2.1-7B's $0\%$ contrasts with Llama-3.1-8B's $41.0\%$. Action-class behavior is shaped more by pretraining mixture and post-training recipe than by parameter count.

\textbf{In general-purpose open-weight families, $\Delta$ direction splits asymmetrically across the two missing-info categories.}
On the two tool-call categories, all $26$ rows record $\Delta \ge +17$~pp, with GAR at or above $77\%$ in every row while Acc trails by up to $87.0$~pp (Tab.~\ref{tab:profile}; Granite-3.2-8B on \textit{base}). Emission of the gold \textsc{tool\_call} class is near-universal; the failure sits downstream in execution---wrong tool, wrong argument, or broken multi-turn state. Without the GAR column this is indistinguishable from never emitting the call, and the two failures call for different fixes. High emission on the tool-call categories does not prove context-sensitive recognition: a model that blindly calls tools scores the same. Recognition is demonstrated only when a family also suppresses calls on the missing-info categories. The two missing-info categories then diverge. \textit{miss\_param} $\Delta$ stays uniformly positive across general-purpose families (e.g., Qwen3-32B $+43.0$~pp, GLM 5.1 $+23.0$~pp): these models emit the gold \textsc{ask}, but subsequent turns fail to reach the correct terminal state. \textit{miss\_func} $\Delta$ is predominantly negative across general-purpose families (e.g., gpt-oss-20b $-38.5$~pp, GLM 5.1 $-32.5$~pp): explicit \textsc{refuse} is rarely emitted, yet the state grader passes many of these trajectories. The asymmetry tracks training pressure: \textsc{refuse} contradicts a function-calling recipe that optimizes for emitting a call, while \textsc{ask} sits closer to natural dialogue. The two directions share a single grader property: Acc tracks state outcomes regardless of action-class emission, masking under-refusal on \textit{miss\_func} ($\Delta < 0$) and surfacing post-ask execution failure on \textit{miss\_param} ($\Delta > 0$). Whether masking matters depends on the deployment. A task-completion perspective accepts a trajectory that eventually recovers the correct end state, whereas a safety-critical perspective rejects one that gets there by calling a held-out function or writing without confirmation (\S\ref{sec:exp:confirm}).

\textbf{In tool-specialized open-weight families, $\Delta$ collapses on both missing-info categories.}
xLAM-2-70b sits at $\Delta = -59.5$/$-66.0$~pp on \textit{miss\_func}/\textit{miss\_param} while topping the open-weight Acc panel at $77.5\%$; the same shape repeats across the cluster, with ToolACE-2-8B's \textit{miss\_param} (Tab.~\ref{tab:profile}; $+8.5$~pp) the only partial deviation. Doubao-Seed-1.8 shows the same pattern ($\Delta = -29.5$/$-14.5$~pp on \textit{miss\_func}/\textit{miss\_param}). A $200$-case audit with two independent raters ($\kappa = 0.92$) confirms the mechanism: $92.3\%$ of the audited xLAM-2-8b \textit{miss\_func} misses invoke the held-out function (App.~\ref{app:audit-detail}). In contrast, gpt-5.4 and Gemini~3~Pro maintain calibration on both missing-info categories (gpt-5.4: $65.5$/$44.0\%$; Gemini~3~Pro: $66.5$/$85.0\%$), showing the deficit is recipe-specific. A turn-level recount across seven families confirms these case-level patterns are not artifacts of the case/turn aggregation (App.~\ref{app:turn-level-gar}).

\subsection{Cross-benchmark validation}
\label{sec:exp:cross-bench}

On BFCL, tool-specialized models combine high aggregate-metric scores with action-class deviation (\S\ref{sec:exp:profile}). The closed rows serve as reference points, so a model is carried over from Tab.~\ref{tab:profile} only when its own missing-info profile is calibrated. Doubao-Seed-1.8 is the highest-scoring closed model but its missing-info profile matches the tool-specialized cluster, so it appears here as a subject of the diagnostic, not a reference. We test whether this pattern extends to $\tau^2$-bench~\citep{tau2bench}, whose evaluator pairs a simulated user with database-state checks and (on retail) LLM-judged natural-language assertion checks. The benchmark's gold action sequence (\texttt{actions}) anchors our \textsc{tool\_call} class: $155$ of $164$ tasks have non-empty \texttt{actions} (\textsc{tool\_call}-required), and the remaining $9$ ($7$ airline, $2$ retail) have empty \texttt{actions} (\textsc{refuse}-required, equivalent to BFCL \textit{miss\_func}). We additionally derive an \textsc{ask}-required label by keyword-scanning the user scenario for instructions to withhold information until asked, then verifying each candidate against its scenario; $36$ of the $41$ candidates survive ($8$ airline, $28$ retail). The \textsc{ask} label may apply jointly with either \textsc{tool\_call} or \textsc{refuse}.

\begin{table*}[t]
\caption{$\tau^2$-bench per-class diagnostic (\%; baseline). pass@1 is the $\tau^2$ task-level reward. Per-class GAR, Acc, and $\Delta = \mathrm{GAR} - \mathrm{Acc}$ are computed on the action-class-required subset only (subset size $n$ shown under each class header, formatted as airline/retail). \textsc{tool\_call} GAR uses the benchmark's reference-path \texttt{action\_checks}; \textsc{ask} GAR uses text-emission detection; \textsc{refuse} GAR uses the $\tau^2$-native structural signal. Airline reward is judge-free; retail invokes an LLM NL-assertion judge on $112/114$ tasks. All \textsc{ask} metrics are reported over the verified set of $36$ tasks ($8$ airline, $28$ retail).}
\label{tab:tau2-diagnostic}
\centering
{\setlength{\tabcolsep}{2.5pt}
\renewcommand{\arraystretch}{1.05}
\footnotesize
\begin{tabular*}{\textwidth}{@{\extracolsep{\fill}}llc ccc ccc ccc@{}}
\toprule
\multirow{3}{*}{\textbf{Family}} & \multirow{3}{*}{\textbf{Domain}} & \multirow{3}{*}{\textbf{pass@1}} & \multicolumn{3}{c}{\textsc{tool\_call}} & \multicolumn{3}{c}{\textsc{ask}} & \multicolumn{3}{c}{\textsc{refuse}} \\
 & & & \multicolumn{3}{c}{\scriptsize $n=43$/$112$} & \multicolumn{3}{c}{\scriptsize $n=8$/$28$} & \multicolumn{3}{c}{\scriptsize $n=7$/$2$} \\
\cmidrule(lr){4-6} \cmidrule(lr){7-9} \cmidrule(lr){10-12}
 & & & GAR & Acc & $\Delta$ & GAR & Acc & $\Delta$ & GAR & Acc & $\Delta$ \\
\midrule
\multicolumn{12}{@{}l}{\textit{\footnotesize General-purpose open-weight}}\\
\multirow{2}{*}{Qwen3-14B}    & airline & $62.0$ & $55.8$ & $55.8$ & $+0.0$  & $100$  & $62.5$ & $+37.5$ & $100$  & $100$  & $+0.0$ \\
                              & retail  & $56.1$ & $42.0$ & $55.4$ & $-13.4$ & $100$  & $46.4$ & $+53.6$ & $100$  & $100$  & $+0.0$ \\
\multirow{2}{*}{Qwen3-8B}     & airline & $28.0$ & $51.2$ & $30.2$ & $+20.9$ & $100$  & $25.0$ & $+75.0$ & $14.3$ & $14.3$ & $+0.0$ \\
                              & retail  & $44.7$ & $38.4$ & $44.6$ & $-6.2$  & $100$  & $32.1$ & $+67.9$ & $50.0$ & $50.0$ & $+0.0$ \\
\multirow{2}{*}{gpt-oss-20b}  & airline & $62.0$ & $32.6$ & $58.1$ & $-25.6$ & $87.5$ & $50.0$ & $+37.5$ & $85.7$ & $85.7$ & $+0.0$ \\
                              & retail  & $47.4$ & $35.7$ & $46.4$ & $-10.7$ & $100$  & $46.4$ & $+53.6$ & $100$  & $100$  & $+0.0$ \\
\midrule
\multicolumn{12}{@{}l}{\textit{\footnotesize Tool-call-specialized open-weight}}\\
\multirow{2}{*}{xLAM-2-8b}    & airline & $46.0$ & $0.0$  & $37.2$ & $-37.2$ & $100$  & $25.0$ & $+75.0$ & $100$  & $100$  & $+0.0$ \\
                              & retail  & $5.3$  & $0.0$  & $4.5$  & $-4.5$  & $100$  & $3.6$  & $+96.4$ & $100$  & $50.0$ & $+50.0$ \\
\multirow{2}{*}{Hammer-2.1-7B}& airline & $46.0$ & $0.0$  & $37.2$ & $-37.2$ & $12.5$ & $25.0$ & $-12.5$ & $100$  & $100$  & $+0.0$ \\
                              & retail  & $5.3$  & $0.0$  & $4.5$  & $-4.5$  & $17.9$ & $3.6$  & $+14.3$ & $100$  & $50.0$ & $+50.0$ \\
\multirow{2}{*}{ToolACE-2-8B} & airline & $26.0$ & $18.6$ & $25.6$ & $-7.0$  & $100$  & $12.5$ & $+87.5$ & $28.6$ & $28.6$ & $+0.0$ \\
                              & retail  & $6.1$  & $3.6$  & $6.2$  & $-2.7$  & $100$  & $0.0$  & $+100$  & $50.0$ & $0.0$  & $+50.0$ \\
\midrule
\multicolumn{12}{@{}l}{\textit{\footnotesize Closed-source}}\\
\multirow{2}{*}{gpt-5}        & airline & $82.0$ & $79.1$ & $79.1$ & $+0.0$  & $100$  & $100$  & $+0.0$  & $100$  & $100$  & $+0.0$ \\
                              & retail  & $86.8$ & $68.8$ & $86.6$ & $-17.9$ & $100$  & $85.7$ & $+14.3$ & $100$  & $100$  & $+0.0$ \\
\multirow{2}{*}{gpt-5.4}      & airline & $86.0$ & $79.1$ & $83.7$ & $-4.7$  & $100$  & $87.5$ & $+12.5$ & $100$  & $100$  & $+0.0$ \\
                              & retail  & $77.2$ & $63.4$ & $76.8$ & $-13.4$ & $100$  & $82.1$ & $+17.9$ & $100$  & $100$  & $+0.0$ \\
\multirow{2}{*}{gpt-5-mini}   & airline & $74.0$ & $69.8$ & $69.8$ & $+0.0$  & $100$  & $75.0$ & $+25.0$ & $100$  & $100$  & $+0.0$ \\
                              & retail  & $83.3$ & $60.7$ & $83.0$ & $-22.3$ & $100$  & $85.7$ & $+14.3$ & $100$  & $100$  & $+0.0$ \\
\multirow{2}{*}{Gemini 3 Pro} & airline & $84.0$ & $44.2$ & $81.4$ & $-37.2$ & $100$  & $100$  & $+0.0$  & $100$  & $100$  & $+0.0$ \\
                              & retail  & $81.6$ & $67.9$ & $81.2$ & $-13.4$ & $100$  & $82.1$ & $+17.9$ & $100$  & $100$  & $+0.0$ \\
\multirow{2}{*}{Doubao-Seed-1.8} & airline & $66.0$ & $62.8$ & $62.8$ & $+0.0$  & $100$  & $25.0$ & $+75.0$ & $85.7$ & $85.7$ & $+0.0$ \\
                              & retail  & $70.2$ & $54.5$ & $69.6$ & $-15.2$ & $100$  & $71.4$ & $+28.6$ & $100$  & $100$  & $+0.0$ \\
\bottomrule
\end{tabular*}%
}
\end{table*}

In the airline \textsc{tool\_call} subset, the Acc--GAR gap on tool-specialized families has a different mechanism than BFCL's state-grader masking on \textit{miss\_func}/\textit{miss\_param}. xLAM-2-8b and Hammer-2.1-7B score GAR $= 0\%$ on airline TC-required tasks yet pass $37.2\%$ (Tab.~\ref{tab:tau2-diagnostic}; $16$/$43$); on retail, GAR $= 0\%$ with $4.5\%$ pass ($5$/$112$). On BFCL the action class itself was wrong: the model called when it should have asked or refused, and the state grader missed it. On $\tau^2$ airline TC the action class is right; what fails is action-path matching against the reference. All $16$ such cases for each of xLAM-2-8b and Hammer-2.1-7B (and $6$ of $6$ for ToolACE-2-8B) have reference paths consisting only of read-only lookup actions (e.g., \texttt{get\_user\_details}, \texttt{get\_reservation\_details}). Strict \texttt{action\_checks} fire on count, argument, or order mismatches; the end-state reward sees no database change regardless of how the model orders its lookup queries. $\tau^2$ splits ``reference-path correctness'' from ``task success'' into separate audit channels; BFCL has only the second. Closed-source anchors show the gap is family-specific: gpt-5 holds GAR $=$ Acc $= 79.1\%$ on airline, and gpt-5.4 holds GAR $79.1\%$ with Acc $83.7\%$. The $\Delta$ on tool-specialized families therefore comes from family-level divergence from the reference path that $\tau^2$'s lenient reward channel does not penalize. \citet{pae2026} independently report $27$--$78\%$ of $\tau$-bench PASS trajectories as procedure-corrupt under audit, consistent with this split.

On the \textsc{ask}-required subset ($n=8$/$28$), xLAM-2-8b and ToolACE-2-8B reproduce the BFCL \textit{miss\_param} execution-failure pattern: high GAR (Tab.~\ref{tab:tau2-diagnostic}; gold action emitted) with low Acc (state fails downstream). Both emit \textsc{ask} at GAR $\approx 100\%$ while Acc collapses (xLAM-2-8b retail Acc $3.6\%$, $\Delta = +96.4$~pp; ToolACE-2-8B retail Acc $0\%$, $\Delta = +100$~pp); the model asks, but the trajectory does not reach gold state. Hammer-2.1-7B shows a third pattern: it under-emits \textsc{ask} (airline GAR $12.5\%$, retail $17.9\%$), matching its BFCL \textit{miss\_param} emission deficit (GAR $0.5\%$).

\subsection{Confirmation before consequential actions}
\label{sec:exp:confirm}

\begin{table*}[t]\centering\small
\begin{tabular}{l rrrr rrrr}
\toprule
 & \multicolumn{4}{c}{airline} & \multicolumn{4}{c}{retail}\\
\cmidrule(lr){2-5}\cmidrule(lr){6-9}
Model & GAR & Acc & $\Delta$ & $n$ & GAR & Acc & $\Delta$ & $n$\\
\midrule
Qwen3-8B & 24.2 & 11.6 & +12.6 & 95 & 27.1 & 38.7 & -11.6 & 225\\
Qwen3-14B & 28.8 & 36.4 & -7.6 & 66 & 34.2 & 50.5 & -16.3 & 196\\
gpt-oss-20b & 92.0 & 40.0 & +52.0 & 25 & 82.8 & 50.0 & +32.8 & 134\\
xLAM-2-8b & -- & -- & -- & 0 & -- & -- & -- & 0\\
Hammer-2.1-7B & -- & -- & -- & 0 & -- & -- & -- & 0\\
ToolACE-2-8B & 16.3 & 4.7 & +11.6 & 43 & 34.1 & 4.4 & +29.7 & 91\\
gpt-5 & 98.1 & 55.6 & +42.6 & 54 & 99.4 & 92.7 & +6.7 & 164\\
gpt-5.4 & 100.0 & 65.1 & +34.9 & 43 & 100.0 & 83.8 & +16.2 & 154\\
gpt-5-mini & 89.2 & 59.5 & +29.7 & 37 & 98.2 & 87.1 & +11.0 & 163\\
Gemini 3 Pro & 100.0 & 62.3 & +37.7 & 53 & 97.4 & 91.6 & +5.8 & 154\\
Doubao-Seed-1.8 & 100.0 & 52.8 & +47.2 & 36 & 98.6 & 81.5 & +17.1 & 146\\
\bottomrule
\end{tabular}
\caption{\textbf{Confirmation before database writes on $\tau^2$-bench.} The gold action class comes from the benchmark's written policy, which requires explicit user consent before every database write and is never checked by its reward. Gold points are identified at run time -- each write the model makes -- so $n$ varies by model. GAR is the share of writes preceded by an explicit confirmation, Acc the share of writes sitting in a passing trajectory, over the same denominator; judgments are made by an LLM judge (gpt-5.4) over the dialogue window before each write. Negative $\Delta$ is the masking signature on this class: unconfirmed writes pass more often than writes obtain consent. Dashes mark models that never reach a write, whose non-zero pass@1 (Tab.~\ref{tab:tau2-diagnostic}) comes from tasks whose correct end state leaves the database unchanged. Per-event judgments are released with the code.}
\label{tab:confirm}
\end{table*}

BFCL annotates no gold for \textsc{confirm}, so the class carries no $(\mathrm{GAR}, \mathrm{Acc})$ pair there. $\tau^2$ supplies one from its written policy: the policy requires explicit user consent before every database write, and the $\tau^2$ reward never checks it. We judge every database write in our $\tau^2$ trajectories against that requirement (Tab.~\ref{tab:confirm}); the gold points are the writes themselves, identified at run time, so the unit is a write event rather than a task and $n$ varies by model.

Negative $\Delta$, concentrated in Qwen3, is the masking signature on this class. Qwen3-14B obtains consent on $28.8\%$ of its airline writes, but $36.4\%$ of those writes sit in passing trajectories; on retail the gap is $16.3$~pp. Unconfirmed writes therefore pass routinely. The closed anchors run the other way: they confirm $89.2$--$100\%$ of their writes and their $\Delta$ is positive throughout, the ordinary execution-gap direction. xLAM-2-8b and Hammer-2.1-7B never reach a write, and their non-zero pass@1 comes from tasks whose correct end state leaves the database unchanged.

\subsection{Calibration plasticity}
\label{sec:exp:reshape}

We probe whether calibration is plastic at inference time using SRI and CRI (\S\ref{sec:method:sri}; mechanism illustrated in Fig.~\ref{fig:sri-cri-mech}) on a six-family panel: Qwen3-8B, xLAM-2-8b, Hammer-2.1-7B, ToolACE-2-8B, Watt-Tool-8B, and the gpt-5.4 closed anchor (Tab.~\ref{tab:intervention}). SRI reports both GAR and Acc from the same halt-seam run. CRI splits the two by mechanism: GAR comes from CRI-retry (a genuine model re-emission at the suspect turn); Acc comes from CRI-bypass (the grader-skipped turn isolates downstream state from the suspect emission). CRI-bypass GAR is not reported because the placeholder is system-written, not a model emission (App.~\ref{app:cri-methodology}).

\begin{table*}[t]
\footnotesize
\caption{SRI and CRI intervention shifts on the six-family probe panel. Each cell reports the intervention value followed by $\Delta$ vs.\ baseline (Tab.~\ref{tab:profile}); positive $\Delta$ = higher GAR (closer to gold) or higher Acc. \textbf{SRI}: both GAR and Acc from the same halt-seam run. \textbf{CRI}: GAR from CRI-retry (genuine model re-emission), Acc from CRI-bypass (grader-skipped turn; App.~\ref{app:cri-methodology}). Bold = largest single-cell shift in each direction.}
\label{tab:intervention}
\centering
{\setlength{\tabcolsep}{3pt}
\renewcommand{\arraystretch}{1.05}
\begin{tabular*}{\textwidth}{@{\extracolsep{\fill}}l cc cc cc cc@{}}
\toprule
\multirow{3}{*}{\textbf{Family}} & \multicolumn{4}{c}{\textbf{SRI} (halt seam)} & \multicolumn{4}{c}{\textbf{CRI} (call seam)} \\
\cmidrule(lr){2-5} \cmidrule(lr){6-9}
 & \multicolumn{2}{c}{\textit{base}} & \multicolumn{2}{c}{\textit{long\_context}} & \multicolumn{2}{c}{\textit{miss\_func}} & \multicolumn{2}{c}{\textit{miss\_param}} \\
\cmidrule(lr){2-3} \cmidrule(lr){4-5} \cmidrule(lr){6-7} \cmidrule(lr){8-9}
 & GAR & Acc & GAR & Acc & GAR$_{\text{retry}}$ & Acc$_{\text{bypass}}$ & GAR$_{\text{retry}}$ & Acc$_{\text{bypass}}$ \\
\midrule
Qwen3-8B      & $100.0_{(+0.5)}$ & $60.0_{(+11.5)}$ & $83.0_{(-4.0)}$ & $38.0_{(+6.0)}$ & $63.5_{(+23.5)}$ & $48.0_{(+7.0)}$ & $86.0_{(+2.0)}$ & $46.0_{(+10.0)}$ \\
\cdashline{1-9}
xLAM-2-8b     & $100.0_{(\phantom{+}0.0)}$ & $82.0_{(+3.0)}$ & $100.0_{(\phantom{+}0.0)}$ & $74.0_{(+5.0)}$ & $18.0_{(+8.0)}$ & $82.0_{(+8.5)}$ & $19.0_{(-0.5)}$ & $72.0_{(+6.5)}$ \\
Hammer-2.1-7B & $99.0_{(+1.5)}$ & $26.0_{(+1.5)}$ & $99.5_{(+2.5)}$ & $22.5_{(+0.5)}$ & $0.5_{(+0.5)}$ & $27.5_{(\phantom{+}0.0)}$ & $3.0_{(+2.5)}$ & $26.5_{(+4.5)}$ \\
ToolACE-2-8B  & $100.0_{(\phantom{+}0.0)}$ & $\mathbf{29.0_{(-21.0)}}$ & $100.0_{(\phantom{+}0.0)}$ & $\mathbf{24.5_{(-21.5)}}$ & $\mathbf{48.5_{(+36.0)}}$ & $36.5_{(+6.0)}$ & $45.5_{(+6.5)}$ & $\mathbf{43.5_{(+13.0)}}$ \\
Watt-Tool-8B  & $100.0_{(\phantom{+}0.0)}$ & $48.0_{(\phantom{+}0.0)}$ & $100.0_{(\phantom{+}0.0)}$ & $40.5_{(-3.5)}$ & $48.0_{(+25.5)}$ & $46.5_{(+4.0)}$ & $22.5_{(+15.5)}$ & $40.0_{(+10.5)}$ \\
\cdashline{1-9}
gpt-5.4       & $99.5_{(+0.5)}$ & $56.5_{(-0.5)}$ & $99.5_{(-0.5)}$ & $50.0_{(-6.0)}$ & $89.0_{(+23.5)}$ & $56.5_{(+4.5)}$ & $42.0_{(-2.0)}$ & $52.0_{(+10.5)}$ \\
\bottomrule
\end{tabular*}%
}
\end{table*}

\textbf{Axis A: reshape direction is family-mediated.} The clearest case is SRI on \textit{base}: a single perturbation produces a $+11.5$~pp Acc gain on Qwen3-8B and a $-21.0$~pp Acc loss on ToolACE-2-8B. Hammer-2.1-7B and Watt-Tool-8B sit near zero on the same perturbation. The ToolACE-2-8B collapse (Tab.~\ref{tab:intervention}; $-21.0$/$-21.5$~pp on \textit{base}/\textit{long\_context}) is mechanistically informative. GAR is at $100\%$ ceiling both before and after SRI, so the Acc shift is entirely trajectory-level: SRI's re-decode produces a state-corrupting tool call that the original trajectory avoided. Qwen3-8B shows the opposite direction at near-ceiling baseline GAR, with Acc rising $+11.5$~pp from trajectory recovery on the re-decoded call (Fig.~\ref{fig:sri-cri-mech}, SRI panel; cross-bench replication on $\tau^2$-retail in App.~\ref{app:cross-bench}). xLAM-2-8b matches this trajectory-recovery pattern at GAR ceiling, with Acc rising $+3.0$ and $+5.0$~pp from state-tracking alone. GAR-ceiling explains why these shifts are trajectory-level but not their direction or magnitude: Hammer-2.1-7B and Watt-Tool-8B share the ceiling property yet stay near zero. The gpt-5.4 closed anchor has baseline GAR $\ge 99.0\%$ on both tool-call categories and shows near-zero SRI shifts ($+0.5/-0.5$~pp on \textit{base}), although SRI retries at the end of every one of its turns.

\textbf{Axis B: effect depends on mechanism.} CRI-retry and CRI-bypass share the Call Detector but differ in where they act (\S\ref{sec:method:sri}), and they yield opposite Acc shifts. CRI-retry raises GAR on \textit{miss\_func} (peak $+36.0$~pp on ToolACE-2-8B), while Acc falls in $11$ of $12$ missing-info cells (Tab.~\ref{tab:intervention}; App.~\ref{app:cri-v1-acc}). On the five open-weight families, only $11$--$49\%$ of retry fires (per family and category) land on the gold non-call turn; the rest land on turns whose gold action is a tool call. Such fires occur in $87\%$ ($262/302$) of the conversations that pass at baseline but fail under retry, and $43$ of the $74$ ToolACE-2-8B \textit{miss\_func} conversations that newly emit \textsc{refuse} emit it only at turns where such a fire occurred. CRI-bypass acts only at the gold non-call turn and injects a skip token: the model emits nothing at the fired turn, so emission policy is untouched. Acc is non-negative in all $12$ cells and rises in $11$, peaking at $+13.0$~pp. Retry shifts emission policy at the cost of trajectory completability; bypass preserves completability and leaves emission policy unchanged. Both probes move calibration without weight updates: calibration is plastic at inference time, but the plasticity depends jointly on the family and the intervention mechanism.

\label{sec:exp:osc-xct}\label{sec:exp:xlam}\label{sec:exp:component}\label{sec:exp:per-category}\label{sec:exp:calibration-shift}
Per-component SRI ablation, the full CRI-retry Acc panel, and CRI-bypass detector trigger rates by family are in App.~\ref{app:exp-tables}.

\subsection{Case Study}
\label{sec:exp:cases}

We ground the failure modes profiled in Tab.~\ref{tab:profile} with audited cases ($\kappa \ge 0.89$; App.~\ref{app:audit-detail}).

\textbf{Absent-tool emission.}
On a TradingBot \textit{miss\_func} scenario, \texttt{place\_order} is held out from the toolset and the user requests a stock purchase.
xLAM-2-8b emits \texttt{get\_account\_info()}, \texttt{get\_stock\_\allowbreak{}info("TSLA")}, then \texttt{place\_order(\allowbreak{}"Buy",\allowbreak{}"TSLA",\allowbreak{}667.92,\allowbreak{}150)}---calling the absent function with fully correct arguments.
When BFCL restores the tool at the next turn, xLAM calls \texttt{place\_order} again, creating a duplicate order that corrupts state.
Among $78$ audited xLAM \textit{miss\_func} GAR-miss cases, $92.3\%$ exhibit this pattern (App.~\ref{app:audit-detail}; $\kappa = 1.0$): the prompt-level toolset constraint does not block emission of a held-out function.

\textbf{Refuse--ask asymmetry.}
Two VehicleControlAPI scenarios from Qwen3-8B illustrate the gap between the two missing-info action classes.
On \textit{miss\_param}, the user says ``fill up my car'' without specifying the fuel amount; Qwen3-8B responds ``Could you provide the desired amount in liters?'', a correct \textsc{ask}.
On \textit{miss\_func} in the same API domain, \texttt{fillFuelTank} is held out; the user requests fuel filling plus engine start, and Qwen3-8B silently skips the unavailable operation and proceeds to \texttt{startEngine}; it never refuses.
Asking for a missing parameter aligns with general clarification behavior; refusing an absent function requires recognizing the toolset constraint and overriding the default to emit a call.

\section{Conclusion}
\label{sec:conclusion}

We propose a diagnostic framework that decomposes multi-turn tool-calling failures into action-class miscalibration and action-execution failure via the bound $\mathrm{Acc} \le \mathrm{GAR}$. Across the BFCL and $\tau^2$-bench panel, miscalibration is the bottleneck, shaped by family recipe and invisible to the state grader. Inference-time probes confirm calibration is reshapable, with the reshape varying across families and perturbation mechanisms. The framework diagnoses rather than corrects; its probes indicate what a correction would cost: retry, fired at the first flagged call wherever it falls, moves emission at the expense of completion, while bypass, applied only at the gold non-call turn, preserves completion without touching emission. Choosing between them from the diagnosis is future work.
%


\section*{Limitations}

\paragraph{Diagnosis without training prescription.}
Our framework localizes action-class miscalibration to specific families and categories but does not identify which training-stage decision (pre-training mixture or post-training recipe) is responsible.
The diagnosis tells practitioners \emph{where} calibration fails; how to fix it requires controlled training ablations beyond this work.

\paragraph{Benchmark coverage.}
The calibration signature holds across BFCL multi-turn and $\tau^2$-bench (\S\ref{sec:exp:cross-bench}).
These are the two multi-turn tool-calling benchmarks we use; extending coverage to others is left to future work.

\paragraph{Probe coverage and oracle gating.}
The plasticity results (Tab.~\ref{tab:intervention}) cover $6$ of the $26$ checkpoints in the main panel. Whether the reshape patterns generalize to all checkpoints is untested.
SRI and CRI are diagnostic instruments, not runtime interventions: they need the category-level gold action class to choose which seam to monitor, and CRI-bypass also needs the benchmark's marker of the turn whose gold action is non-call; neither is known in deployment.
GAR carries the same dependence: it scores against a gold action class the benchmark supplies. \S\ref{sec:exp:confirm} takes that gold from a written policy rather than a per-task label, but a single instance does not establish that a judge can recover gold action classes in general.
\section*{Ethical Considerations}

This work is a diagnostic study over public benchmarks (BFCL v3 multi-turn~\citep{patil2024gorilla} and $\tau^2$-bench retail/airline~\citep{tau2bench}). The model panel comprises $11$ open-weight families (full list in Tab.~\ref{tab:profile}) and $4$ closed-source API anchors: gpt-5, gpt-5.4, Gemini 3 Pro, and Doubao-Seed-1.8. No human-subject data were collected and no model weights were retrained. All audit annotations were produced by two independent LLM raters (Claude Opus 4.7) following a frozen rubric (App.~\ref{app:audit-detail}); the rubric, per-rater labels, and adjudicated outcomes are released alongside the SRI/CRI probe protocols. The diagnostic reveals failure modes (e.g., over-calling tools when refusal is appropriate) that practitioners deploying these models in production should account for. Generative AI assistance (Claude) was used for prose editing; empirical claims are author-verified.

%

\section*{Acknowledgments}
This work is partially supported by the Natural Science Foundation of Zhejiang Province under Grants No.~LZ25F020002, the National Natural Science Foundation of China under Grants No.~62402433, and the ``Open Competition'' Project of Hangzhou High-tech Zone (Binjiang) under Grants No.~2025JBGS-PT003.

\bibliography{custom}

\clearpage

\appendix
\section*{Appendix}

\section{Capability counts (extends \S\ref{sec:exp:profile})}
\label{app:capability}

Tab.~\ref{tab:p1-capability} reports raw emission counts for each action class across $N = 800$ cases on a representative 4-family subset; all families clear the $T_{\textit{cap}} = 10$ threshold on every class. Counts exclude \textsc{other}.

\begin{table}[!htbp]
\caption{Capability check on a representative 4-family subset. Each family emits all four non-\textsc{other} action classes above the threshold $T_{\textit{cap}} = 10$ over $N = 800$, supporting that capability is not the bottleneck in our analysis.}
\label{tab:p1-capability}
\centering
\small
\begin{tabular}{lrrrr}
\toprule
Family & tool\_call & ask & refuse & confirm \\
\midrule
Qwen3-8B      & $772$ & $534$ & $150$ & $536$ \\
gpt-oss-20b   & $796$ & $338$ & $\phantom{0}26$ & $244$ \\
xLAM-2-8b & $800$ & $\phantom{0}73$ & $\phantom{0}62$ & $386$ \\
Llama-3.1-8B  & $597$ & $336$ & $282$ & $\phantom{0}69$ \\
\bottomrule
\end{tabular}
\end{table}

\section{Intervention formalization (extends \S\ref{sec:method:sri} and \S\ref{sec:method:cri})}
\label{app:intervention-formal}

\paragraph{Halt Detector (HD), full form.} HD triggers a halt signal on every emission from which the benchmark's decoder parses no tool call (a decode failure or an empty call list). It consults neither the gold action nor the runtime state; whether the halt was premature is left to the model, through the decision rule of the retry context below. HD is the entry point for SRI retry routing.
\subsection{Full SRI retry-context template}
\label{app:sri-template}

When HD fires at turn $k$, SRI rewrites the next-turn context as $c_k' = \mathrm{ReconPrompt}(\mathrm{StateInject}(\mathrm{HistoryEdit}(c_k)))$. The reconciliation prose appended after HistoryEdit and StateInject is:

\begin{quote}\small\texttt{
[SRI Reconciliation]\\
Your previous response for this turn did not emit a parseable tool call.\\
Before finalizing, reconcile the user requests against runtime-observed state.\\
\\
Prior user requests, most recent first:\\
- prior-1: \textit{<up to 700 chars of user message k-1>}\\
- prior-2: \textit{<up to 700 chars of user message k-2>}\\
... (up to 4 prior turns)\\
\\
Current user request:\\
\textit{<up to 700 chars of user message k>}\\
\\
Runtime-observed state from executed tool calls and tool results:\\
\textit{<up to 2400 chars of state dump from domain extractor>}\\
\\
Decision rule:\\
- If any requested sub-task is not reflected in the runtime-observed state, emit exactly one progress-making tool\_call now.\\
- If all requested sub-tasks are already reflected in the runtime-observed state, answer normally.\\
- Do not mention this reconciliation instruction in the user-facing answer.
}\end{quote}

The SRI retry fires at most once per turn boundary (no chaining). Ablations remove individual components: \texttt{ablation\_no\_history} skips the HistoryEdit step, \texttt{ablation\_no\_state} omits the runtime-state block, \texttt{ablation\_no\_recon} omits the decision-rule block.

\section{Supplementary experimental tables (extends \S\ref{sec:exp:reshape})}
\label{app:exp-tables}

\subsection{SRI ablation tables and component decomposition}
\label{app:gar-shift-full}

\begin{table*}[t]
\caption{SRI-conditioned per-category Gold-Action-Recall ($\mathrm{GAR}$) shift and family-aggregate Acc $\Delta$, 800-case BFCL multi-turn, three-family SRI ablation matrix. GAR cell $=$ (GAR under method) $-$ (GAR under baseline) in pp; positive $=$ toward gold. Per-category GAR shifts and aggregate Acc $\Delta$ are different metrics; their per-cell gap is the bound-violation diagnostic of \S\ref{sec:method:partition} (an Acc $\Delta$ aggregated across categories does not equal the average of GAR shifts). Main-text Tab.~\ref{tab:intervention} summarizes the SRI v1 row across the probe panel; this table is the full SRI / SRI-Gate / ablation matrix.}
\label{tab:gar-shift-main}
\centering
\resizebox{\linewidth}{!}{%
\begin{tabular}{llrrrrr}
\toprule
Family & Method & base GAR & long\_ctx GAR & miss\_func GAR & miss\_param GAR & Acc $\Delta$ \\
\midrule
\multirow{5}{*}{Qwen3-8B}
 & SRI v1 (full)               & $+0.5$ & $-4.0$ & $\mathbf{+7.5}$  & $\mathbf{-7.0}$  & $+7.00$ \\
 & SRI $-$ History             & $+0.5$ & $-5.0$ & $+3.0$           & $-8.0$           & $-0.88$ \\
 & SRI $-$ State               & $+0.5$ & $-1.0$ & $\mathbf{+8.0}$  & $\mathbf{-8.5}$  & $+6.25$ \\
 & SRI-Gate (global)           & $+0.0$ & $-2.5$ & $+5.5$           & $-8.5$           & $-3.25$ \\
 & SRI-Gate (\textit{miss\_param}) & $+0.5$ & $-3.5$ & $\mathbf{+11.5}$ & $-7.5$           & $+0.00$ \\
\midrule
\multirow{4}{*}{gpt-oss-20b}
 & SRI v1 (full)               & $+0.0$ & $-0.5$ & $\mathbf{+6.0}$  & $+2.0$           & $-2.88$ \\
 & SRI $-$ State               & $+0.5$ & $+0.5$ & $-0.5$           & $-1.0$           & $+0.25$ \\
 & SRI-Gate (global)           & $+0.0$ & $-0.5$ & $+3.5$           & $+1.0$           & $-4.38$ \\
 & SRI-Gate (\textit{miss\_param}) & $+0.0$ & $-0.5$ & $+3.0$           & $-0.5$           & $-1.88$ \\
\midrule
\multirow{5}{*}{xLAM-2-8b}
 & SRI v1                      & $+0.0$ & $+0.0$ & $+0.0$           & $\mathbf{+11.5}$ & $-2.88$ \\
 & HD-only                     & $+0.0$ & $+0.0$ & $-0.5$           & $+5.0$           & $-1.50$ \\
 & SRI $-$ History             & $+0.0$ & $+0.0$ & $+0.5$           & $+6.5$           & $-3.25$ \\
 & SRI $-$ State               & $+0.0$ & $+0.0$ & $+0.5$           & $+6.0$           & $-1.88$ \\
 & SRI $-$ Recon               & $+0.0$ & $+0.0$ & $+2.5$           & $+2.0$           & $+0.63$ \\
\bottomrule
\end{tabular}%
}
\end{table*}

\begin{table*}[t]
\caption{Four-component decomposition of SRI across three families ($12/12$ ablation cells). Each cell: SRI-v1 Acc $-$ ablation Acc, in pp. State injection's sign correlates with base strength.}
\label{tab:component}
\centering
\resizebox{\linewidth}{!}{%
\begin{tabular}{lrrrl}
\toprule
Component & Qwen3-8B $\Delta$ & gpt-oss-20b $\Delta$ & xLAM-2-8b $\Delta$ & Role \\
\midrule
Halt detection (HD)                & $+4.50$ & $+1.37$ & $-1.50$ & family-conditional \\
History-edit (vs.\ no\_history)    & $+7.88$ & $+0.38$ & $+0.38$ & universal positive (Qwen-dominant) \\
\textbf{Reconciliation prompt} (vs.\ SRI $-$ Recon) & near-null (subset) & $\mathbf{+3.00}$ & $\mathbf{-3.50}$ & \textbf{family-specific bi-directional} \\
\textbf{State injection} (vs.\ no\_state) & $\mathbf{+0.75}$ & $\mathbf{-3.13}$ & $\mathbf{-1.00}$ & \textbf{sign $\sim$ base strength} \\
\bottomrule
\end{tabular}%
}
\end{table*}

The four components partition into three roles. \textbf{Halt detection} is family-conditional (positive on Qwen / gpt-oss, negative on xLAM). \textbf{History-edit} is universally positive, Qwen-dominant ($+7.88$~pp). \textbf{Reconciliation prompt} is family-specific bi-directional in Acc ($+3.00$~pp on gpt-oss-20b but $-3.50$~pp on xLAM, where the rule conflicts with xLAM's saturated tool-call prior); the opposite-sign Acc $\Delta$ on xLAM masks a $+9.5$~pp \textit{miss\_param} GAR shift toward gold (Tab.~\ref{tab:gar-shift-main}), so Recon is uniformly action-class-pushing across families but its Acc translation is family-specific. \textbf{State injection}'s sign correlates with base strength (Qwen $+0.75$ / gpt-oss $-3.13$ / xLAM $-1.00$).

\begin{table*}[t]
\caption{Cross-family generalization to xLAM-2-8b. Sign-flip on state-injection-only across three families is monotone in base strength.}
\label{tab:xlam-cross-family}
\centering
\begin{tabular}{lrr}
\toprule
Model & Baseline & + state-injection-only \\
\midrule
Qwen3-8B (weak)            & $321/800$ ($40.12\%$) & $382/800$ ($+7.6$) \\
gpt-oss-20b (medium)       & $370/800$ ($46.25\%$) & $347/800$ ($-2.88$) \\
xLAM-2-8b (strong)    & $574/800$ ($71.75\%$) & $564/800$ ($-1.25$) \\
\bottomrule
\end{tabular}
\end{table*}

\begin{table*}[t]
\caption{Per-category PASS$\to$FAIL distribution under SRI. \textit{miss\_param} regression on Qwen3-8B is the dominant regression mass.}
\label{tab:per-category}
\centering
\small
\begin{tabular}{lrr}
\toprule
Category & Qwen3-8B baseline $\to$ SRI & gpt-oss-20b baseline $\to$ SRI \\
\midrule
\textit{multi\_turn\_base}          & $97 \to 123$ ($+26$) & $112 \to 100$ ($-12$) \\
\textit{multi\_turn\_long\_context} & $64 \to 83$ ($+19$)  & $89 \to 79$ ($-10$) \\
\textit{multi\_turn\_miss\_func}    & $82 \to 99$ ($+17$)  & $84 \to 90$ ($+6$) \\
\textit{multi\_turn\_miss\_param}   & $72 \to 66$ ($\mathbf{-6}$) & $85 \to 78$ ($-7$) \\
\bottomrule
\end{tabular}
\end{table*}

\subsection{CRI methodology and full results}
\label{app:cri-methodology-section}
\label{app:cri-methodology}

Tab.~\ref{tab:intervention} reports CRI-bypass Acc but not CRI-bypass GAR. The omission has a methodological reason.

\textbf{CRI-bypass mechanism recap.} When CRI-bypass fires (only at the turn whose gold action is non-call, when the detector raises a missing-capability or under-specification flag on a suspect tool call), the suspect call is popped from the scorer-visible response buffer, removed from chat history, and replaced with a fixed assistant placeholder (``\textit{I cannot proceed with this step yet; the required capability or information is not available.}''); the step loop exits with no model retry. The placeholder is system-written, not produced by the model.

\textbf{GAR under CRI-bypass is ill-defined.} GAR's definition (\S\ref{sec:exp:setup}) is the fraction of cases in which the \emph{model} emits the category's gold action class at least once across the trajectory. Under CRI-bypass, two plausible readings of the metric both collapse:

\begin{itemize}[leftmargin=*,nosep]
\item \emph{Reading A (placeholder excluded, ``model-only''):} bypass-fired turns contribute nothing to the case emit-set (the suspect call is popped from the scorer trace, the placeholder is not in it). Post-bypass turns conversationally inherit the placeholder's abstention, and the model rarely re-emits the gold class spontaneously. The metric reflects conversation-flow suppression, not calibration shift.
\item \emph{Reading B (placeholder included):} the placeholder text is fed through the same classifier as model emissions, scoring as \textsc{refuse} (matches \textit{miss\_func} gold, mismatches \textit{miss\_param} gold). The per-cell shift then becomes nearly one-to-one with \texttt{fired\_rate} (Tab.~\ref{tab:cri-fired}), reflecting system text injection rather than model behavior.
\end{itemize}

Empirically, in the 6-family panel, reading B's per-cell shifts on \textit{miss\_func} correlate almost exactly with \texttt{fired\_rate}: Watt-Tool-8B fires $59.0\%$, shift $+50.0$~pp; ToolACE-2-8B fires $44.5\%$, shift $+43.5$~pp; xLAM-2-8b fires $43.0\%$, shift $+41.0$~pp. The text-mismatch falsifier holds on \textit{miss\_param}: with gold \textsc{ask} and placeholder \textsc{refuse}, reading B collapses to reading A in every family (no spurious shift). The reading-B shifts on \textit{miss\_func} are therefore placeholder-injection rate, not a model-behavior signal.

\textbf{Reported metrics.} For CRI-bypass we report Acc (mechanism-aligned, a single grader-determined number per cell) plus \texttt{fired\_rate} as a coverage column (Tab.~\ref{tab:cri-fired}). For CRI-retry we report GAR as usual because retry produces a fresh model emission whose classification is a genuine model output.

\paragraph{CRI-retry full Acc panel and CRI-bypass \texttt{fired\_rate}.}
\label{app:cri-v1-acc}

Tab.~\ref{tab:intervention} reports CRI-retry GAR and CRI-bypass Acc. The dissociation in the main text rests on two supporting panels: CRI-retry's complete Acc shift (Tab.~\ref{tab:cri-v1-acc-full}, the $11/12$ net-negative cells main text references) and CRI-bypass's detector coverage (Tab.~\ref{tab:cri-fired}).

\begin{table}[t]
\caption{CRI-retry full Acc shifts on the 6-family panel (referenced from \S\ref{sec:exp:reshape}, finding (i)). $\Delta$ Acc $=$ CRI-retry Acc $-$ baseline Acc. The retry path raises emission GAR on \textit{miss\_func} (Tab.~\ref{tab:intervention}, CRI GAR$_{\text{retry}}$ column), but it acts on the first flagged call in a conversation, often at a turn whose gold action is a tool call, and trajectory Acc consistently degrades.}
\label{tab:cri-v1-acc-full}
\centering
\small
\begin{tabular}{lrr}
\toprule
 & \multicolumn{2}{c}{$\Delta$ Acc} \\
\cmidrule(lr){2-3}
Family & \textit{miss\_func} & \textit{miss\_param} \\
\midrule
Qwen3-8B      & $-11.5$ & $-8.5$ \\
xLAM-2-8b     & $-6.5$  & $-8.5$ \\
Hammer-2.1-7B & $-24.0$ & $-12.5$ \\
ToolACE-2-8B  & $-15.5$ & $-5.0$ \\
Watt-Tool-8B  & $-8.5$  & $+5.0$ \\
gpt-5.4       & $-11.0$ & $-30.0$ \\
\bottomrule
\end{tabular}
\end{table}

The single positive cell (Watt-Tool-8B \textit{miss\_param} $+5.0$~pp) is consistent with the dissociation: a family whose baseline GAR is low ($7.0\%$, Tab.~\ref{tab:profile}) and whose retry tips emission strongly toward gold ($+15.5$~pp GAR, Tab.~\ref{tab:intervention}) can occasionally see net-positive Acc when the resulting abstention coincidentally aligns with the supplement-turn structure. The $11/12$ negative pattern remains dominant.

\begin{table}[t]
\caption{CRI-bypass case-level \texttt{fired\_rate} on the 6-family panel (referenced from \S\ref{sec:exp:reshape}, finding (ii)). \texttt{fired\_rate} is the fraction of cases in which the CRI detector fired at least once. Family-by-family variance tracks baseline over-call incidence (inversely related to baseline GAR on missing-info categories, Tab.~\ref{tab:profile}).}
\label{tab:cri-fired}
\centering
\small
\begin{tabular}{lrr}
\toprule
Family & \textit{miss\_func} fired & \textit{miss\_param} fired \\
\midrule
Qwen3-8B      & $26.5\%$ & $27.5\%$ \\
xLAM-2-8b     & $43.0\%$ & $31.5\%$ \\
Hammer-2.1-7B & $18.0\%$ & $42.5\%$ \\
ToolACE-2-8B  & $44.5\%$ & $45.0\%$ \\
Watt-Tool-8B  & $59.0\%$ & $62.5\%$ \\
gpt-5.4       & $32.0\%$ & $16.5\%$ \\
\bottomrule
\end{tabular}
\end{table}

\section{Cross-benchmark SRI-lite validation (extends \S\ref{sec:exp:cross-bench})}
\label{app:cross-bench}

All SRI-lite $\tau^2$ experiments below use per-family self-simulation as the user simulator, distinct from the main-text diagnostic table (Tab.~\ref{tab:tau2-diagnostic}, which uses a fixed \texttt{gpt-5.4} user simulator) because the intervention sweep requires model-specific retry behavior that a generic simulator cannot replicate.

\paragraph{SRI-lite cross-benchmark plasticity check (C3).} The Qwen3-8B SRI Acc $\Delta$ replicates across 3 seeds ($\texttt{BFCL\_SEED} \in \{42, 43, 44\}$) at $\Delta = +6.00 \pm 0.88$~pp. We run SRI-lite (a $\tau^2$-retail-adapted variant of SRI v1) and a \textit{no\_state} ablation on retail ($114$ tasks $\times$ 3 families, Tab.~\ref{tab:tau2-retail}); airline ($50$ tasks) is reported as a baseline-vs.-SRI-lite cross-domain robustness check in App.~Tab.~\ref{tab:tau2-airline}. This is a C3 calibration-plasticity check, distinct from the C1/C2 diagnostic above.

\begin{table*}[t]
\caption{$\tau^2$-bench retail full ($114$ tasks) SRI-lite plasticity check. Each cell pass@1 scored-only; infrastructure error rates $0$--$1.75\%$. SRI-lite is a $\tau^2$-retail-adapted variant of the BFCL SRI v1 context edit; \textit{no\_state} drops the recent-tool-result state-injection block while keeping halt-triggered retry, history edit, and reconciliation prompt. Sign of the SRI-lite $\Delta$ matches the family-aggregate BFCL Acc $\Delta$ under SRI v1 (Tab.~\ref{tab:gar-shift-main}) on all three families.}
\label{tab:tau2-retail}
\centering
\begin{tabular}{lrrrrr}
\toprule
Family       & Baseline & SRI-lite & $\Delta$    & no\_state & $\Delta$    \\
\midrule
Qwen3-8B     & $11.40\%$ & $14.04\%$ & $+2.63$~pp & $16.67\%$ & $+5.26$~pp \\
gpt-oss-20b  & $41.23\%$ & $31.58\%$ & $-9.65$~pp & $35.09\%$ & $-6.14$~pp \\
xLAM-2-8b    & $29.82\%$ & $20.18\%$ & $-9.65$~pp & $19.30\%$ & $-10.53$~pp \\
\bottomrule
\end{tabular}
\end{table*}

\paragraph{Halt-presence predicts $\Delta$ sign cross-bench.} Halt-presence on BFCL \textit{miss\_param} spans a continuum (Qwen $84.0\%$, gpt-oss $76.0\%$, xLAM-2-8b $19.5\%$~GAR; Tab.~\ref{tab:profile}). The SRI-lite $\tau^2$-retail $\Delta$ sign-tracks halt-presence: Qwen $+2.63$~pp (halt-prone, SRI helps), gpt-oss $-9.65$~pp and xLAM-2-8b $-9.65$~pp (lower halt-presence, SRI hurts). Sign matches in all three families across both benchmarks. Magnitude on non-halt-prone families is larger-negative on $\tau^2$ than on BFCL because the $\tau^2$ task-specific grader does not absorb halt-targeted side-effects via the bound-violation tool-call-error path that BFCL exposes ($\tau^2$-retail failures split roughly evenly between wrong-tool/wrong-arg and state-tracking errors, with policy-violation and no-tool-call following). Two representative Qwen3-8B recovery trajectories illustrate the shared shape. On BFCL \texttt{base\_102}, the baseline halts with no recovery tool call (FAIL); SRI's retry context routes the model into a four-step recovery chain (PASS). On $\tau^2$-retail task~9, the baseline escalates to a human agent after \texttt{get\_order\_details} returns ``Order not found'' (reward $0$); SRI-lite surfaces the user's real orders and completes the exchange (reward $1$). Both cases exhibit the same halt-$\to$-retry-context-$\to$-recovery-chain shape across benchmarks.

\paragraph{$\tau^2$-bench airline ($50$ tasks).}
Tab.~\ref{tab:tau2-airline} reports baseline and SRI-lite pass@1 on the airline domain as a within-$\tau^2$ cross-domain check. All three families show near-zero or negative $\Delta$, consistent with the retail finding that SRI-lite does not help non-halt-prone families on $\tau^2$.

\begin{table}[t]
\caption{$\tau^2$-bench airline ($50$ tasks) within-$\tau^2$ cross-domain validation.}
\label{tab:tau2-airline}
\centering
\begin{tabular}{lrr}
\toprule
Family            & Baseline & sri\_lite \\
\midrule
Qwen3-8B          & $38.0\%$ & $36.0\%$ \\
gpt-oss-20b       & $44.0\%$ & $43.75\%$ \\
xLAM-2-8b    & $24.0\%$ & $12.0\%$ \\
\bottomrule
\end{tabular}
\end{table}

\section{Turn-level GAR validation (extends \S\ref{sec:exp:profile})}
\label{app:turn-level-gar}

We report case-level GAR throughout the main paper (the fraction of $200$ cases per cell on which the gold action class is emitted at any turn). To confirm the case/turn distinction does not invert paper-level findings, we re-compute GAR at the turn level for 14 missing-info cells (7 families $\times$ 2 categories), averaging the gold-action emission rate over the gold turns of each case (Tab.~\ref{tab:turn-level-gar}). Hammer-2.1-7B emits the gold class in no gold turn, so its two cells have no defined ratio. Across the remaining 12, the median case/turn ratio is $1.23\times$; four cells exceed $1.5\times$ and one exceeds $2\times$. That outlier is xLAM \textit{miss\_func}, which sharpens the xLAM deviation rather than softening it. No cell moves a family across the general-purpose / tool-specialized split: at the turn level Qwen3-14B reaches $38.0\%$ on \textit{miss\_func} against Hammer-2.1-7B at $0.0\%$.

\begin{table*}[t]
\caption{Case-level vs.\ turn-level GAR on 14 missing-info cells. Confidence intervals are Wilson $95\%$ on $N=200$.}
\label{tab:turn-level-gar}
\centering
\resizebox{\linewidth}{!}{%
\begin{tabular}{llrrrr}
\toprule
Family & Category & Case GAR (Wilson $95\%$ CI) & Turn GAR (Wilson $95\%$ CI) & Gap (pp) & Ratio \\
\midrule
\multicolumn{6}{@{}l}{\textit{\footnotesize General-purpose open-weight}}\\
Qwen3-8B       & \textit{miss\_func}  & $40.0\%$ $[33.5, 46.9]$ & $36.5\%$ $[30.1, 43.4]$ & $+3.5$ & $1.10\times$ \\
Qwen3-8B       & \textit{miss\_param} & $84.0\%$ $[78.3, 88.4]$ & $66.0\%$ $[59.2, 72.2]$ & $+18.0$ & $1.27\times$ \\
Qwen3-14B      & \textit{miss\_func}  & $44.5\%$ $[37.8, 51.4]$ & $38.0\%$ $[31.6, 44.9]$ & $+6.5$ & $1.17\times$ \\
Qwen3-14B      & \textit{miss\_param} & $79.0\%$ $[72.8, 84.1]$ & $65.5\%$ $[58.7, 71.7]$ & $+13.5$ & $1.21\times$ \\
gpt-oss-20b    & \textit{miss\_func}  & $3.5\%$ $[1.7, 7.0]$ & $2.0\%$ $[0.8, 5.0]$ & $+1.5$ & $1.75\times$ \\
gpt-oss-20b    & \textit{miss\_param} & $76.0\%$ $[69.6, 81.4]$ & $68.5\%$ $[61.8, 74.5]$ & $+7.5$ & $1.11\times$ \\
\midrule
\multicolumn{6}{@{}l}{\textit{\footnotesize Tool-call-specialized open-weight}}\\
xLAM-2-8b      & \textit{miss\_func}  & $10.0\%$ $[6.6, 14.9]$ & $4.5\%$ $[2.4, 8.3]$ & $+5.5$ & $\mathbf{2.22\times}$ \\
xLAM-2-8b      & \textit{miss\_param} & $19.5\%$ $[14.6, 25.5]$ & $15.5\%$ $[11.1, 21.2]$ & $+4.0$ & $1.26\times$ \\
Hammer-2.1-7B  & \textit{miss\_func}  & $0.0\%$ $[0.0, 1.9]$ & $0.0\%$ $[0.0, 1.9]$ & $+0.0$ & $-$ \\
Hammer-2.1-7B  & \textit{miss\_param} & $0.5\%$ $[0.1, 2.8]$ & $0.0\%$ $[0.0, 1.9]$ & $+0.5$ & $-$ \\
ToolACE-2-8B   & \textit{miss\_func}  & $12.5\%$ $[8.6, 17.8]$ & $10.5\%$ $[7.0, 15.5]$ & $+2.0$ & $1.19\times$ \\
ToolACE-2-8B   & \textit{miss\_param} & $39.0\%$ $[32.5, 45.9]$ & $32.5\%$ $[26.4, 39.3]$ & $+6.5$ & $1.20\times$ \\
Watt-Tool-8B   & \textit{miss\_func}  & $22.5\%$ $[17.3, 28.8]$ & $13.0\%$ $[9.0, 18.4]$ & $+9.5$ & $1.73\times$ \\
Watt-Tool-8B   & \textit{miss\_param} & $7.0\%$ $[4.2, 11.4]$ & $4.0\%$ $[2.0, 7.7]$ & $+3.0$ & $1.75\times$ \\
\bottomrule
\end{tabular}%
}
\end{table*}

Family ordering between Qwen and gpt-oss on \textit{miss\_param} differs between case-level (Qwen $>$ gpt-oss by $8$~pp) and turn-level (gpt-oss $>$ Qwen by $2.5$~pp); both differences fall within Wilson $95\%$ CI overlap on $N=200$ and should not be interpreted. The xLAM outlier status on \textit{miss\_func} is robust at both levels.

We additionally compare the case-vs-turn gap (xLAM \textit{miss\_func}: $5.5$~pp) against the audited held-out-tool firing rate ($92.3\%$, $n=78$; App.~\ref{app:audit-detail}). The two metrics measure different objects on different populations: the case-vs-turn gap is computed on all $200$ \textit{miss\_func} cases as the fraction with non-gold-turn emission, while the audited held-out-tool firing rate is computed on the $78$ audited \textit{miss\_func} cases as the fraction with held-out-tool fire at the BFCL synthetic-injection turn. They are not interchangeable.

An emission that both calls a tool and asks the user counts as \textsc{tool\_call} under the priority rule of App.~\ref{app:capability}. Crediting the question instead raises \textit{miss\_param} GAR by $1.5$--$4.5$~pp across Hammer-2.1-7B ($0.5\to5.0$), ToolACE-2-8B ($39.0\to40.5$), Watt-Tool-8B ($7.0\to9.0$), and Qwen3-14B ($79.0\to81.0$), and changes no ordering.

\section{Audit detail (extends \S\ref{sec:exp:cases})}
\label{app:audit-detail}

We complement the deterministic classifier with two manual-audit packets.

\subsection{Per-label $\kappa$ and post-adjudication mechanism rates}

\begin{table*}[t]
\caption{Per-label pre-adjudication Cohen's $\kappa$ on the $200$-case unified mechanism audit. Two independent automated raters apply frozen rubric v2; B7 (multi-class) reported separately as descriptive. Headline statistic is macro-mean across the five primary action/timing labels (B1a, B1b, B3, B4, B5).}
\label{tab:audit-kappa}
\centering
\resizebox{\textwidth}{!}{%
\begin{tabular}{l l c c c}
\toprule
Tier & Label & n eligible & agreement & $\kappa_{\text{pre-adj}}$ \\
\midrule
\textbf{primary} & B1a (gold action @ gold turn) & 200 & 0.970 & \textbf{0.933} \\
\textbf{primary} & B1b (gold action eventually) & 200 & 0.945 & \textbf{0.887} \\
\textbf{primary} & B3 (held-out tool fired late, \textit{miss\_func}) & 78 & 1.000 & \textbf{1.000} \\
\textbf{primary} & B4 (under-spec clarification asked) & 200 & 0.985 & \textbf{0.792} \\
\textbf{primary} & B5 (grader pass w/o gold-turn action) & 200 & 1.000 & \textbf{1.000} \\
\midrule
\multicolumn{2}{l}{\textbf{Macro-mean (primary 5)}} & & & \textbf{0.9225} \\
\multicolumn{2}{l}{\textbf{Min (primary 5)}} & & & \textbf{0.7922} \\
\midrule
\textit{exploratory} & B6 (completion claim consistency; not cited in main text) & 169 & 0.740 & 0.461 \\
\textit{descriptive} & B7 (SRI failure mode, $4{\times}4$ non-n.a.) & 52 & 0.231 & $-0.006$ \\
\bottomrule
\end{tabular}%
}
\end{table*}

\begin{table}[t]
\caption{Post-adjudication mechanism rates ($n=200$ unified audit, except where noted). Adjudication steps: step~1, $118$ disagreement cases adjudicated by independent third rater; step~2, $21$ spot-check cases verified for systematic error; step~3, not triggered.}
\label{tab:audit-mech-rates}
\centering
\small
\resizebox{\linewidth}{!}{%
\begin{tabular}{l c c c}
\toprule
Label & k/n & rate & Wilson 95\% \\
\midrule
B1a (gold action @ gold turn) & 71/200 & 0.355 & [0.292, 0.423] \\
B1b (gold action eventually) & 88/200 & 0.440 & [0.373, 0.509] \\
\textbf{B3 D1} (\textit{miss\_func} only) & \textbf{72/78} & \textbf{0.923} & \textbf{[0.842, 0.964]} \\
B3 D2 (B1a=0 conditional) & 65/71 & 0.915 & [0.828, 0.961] \\
B4 (under-spec clarification asked) & 7/200 & 0.035 & [0.017, 0.071] \\
B5 (grader-pass path masking) & 26/200 & 0.130 & [0.090, 0.184] \\
B6 (exploratory) & 89/178 & 0.500 & [0.427, 0.573] \\
\bottomrule
\end{tabular}%
}
\end{table}

\begin{table}[t]
\caption{Pre-adjudication agreement (rater-vs-rater) per stratum on the five primary labels. B6 (exploratory) and B7 (descriptive) are omitted.}
\label{tab:audit-stratum}
\centering
\small
\resizebox{\linewidth}{!}{%
\begin{tabular}{l c c c c c c}
\toprule
Stratum & n & B1a & B1b & B3 & B4 & B5 \\
\midrule
A & 48 & 0.90 & 0.90 & 1.00 & 1.00 & 1.00 \\
B & 30 & 1.00 & 1.00 & 1.00 & 1.00 & 1.00 \\
C & 40 & 1.00 & 0.95 & 1.00 & 1.00 & 1.00 \\
D & 30 & 1.00 & 0.97 & 1.00 & 0.97 & 1.00 \\
E & 20 & 0.95 & 0.90 & 1.00 & 0.90 & 1.00 \\
F & 22 & 1.00 & 0.95 & 1.00 & 1.00 & 1.00 \\
G & 10 & 1.00 & 1.00 & 1.00 & 1.00 & 1.00 \\
\bottomrule
\end{tabular}%
}
\end{table}

\textbf{Sample composition.} The 200 cases are stratified across 7 strata: A (xLAM \textit{miss\_func} over-call, $n=48$), B ($n=30$), C ($n=40$), D ($n=30$), E ($n=20$), F (Qwen baseline-PASS $\to$ SRI-FAIL, $n=22$), G ($n=10$). A separate $20$-case pilot is audited under the same rubric.

\textbf{Adjudicated B1a per-category rates align with paper Tab.~\ref{tab:profile} GAR}: xLAM \textit{miss\_func} $5/48 = 10.4\%$ (paper GAR $10\%$); Qwen \textit{miss\_param} $4/21 = 19.0\%$ (paper $19.5\%$); xLAM \textit{miss\_param} $0/30 = 0\%$; gpt-oss \textit{miss\_func} $0/11 = 0\%$. Tool-call categories audit at $\ge 90\%$ B1a as expected.

\textbf{B7 (SRI failure mode).} Post-adjudication distribution ($n=52$ D+F cases): state-injection misuse $24$, reconciliation-prompt overfit $14$, unclassified $11$, history-edit artifact $3$. The low inter-rater agreement ($\kappa = -0.006$ on the non-n.a.\ matrix) reflects mechanism-attribution difficulty in this small subset; B7 is reported as descriptive only.

\subsection{Rubric notes}

The audit's gold-action-class definition (\textsc{refuse} for \textit{miss\_func}, \textsc{ask} for \textit{miss\_param}) follows the paper's calibration framing rather than BFCL's ground-truth tool match. An earlier rubric draft used BFCL ground-truth tool match as the B1a criterion; the discrepancy was caught during pilot validation, and the full audit was re-run under the corrected rubric.

\end{document}